\pdfoutput=1
\documentclass[11pt]{article}

\usepackage[T1]{fontenc}
\usepackage[utf8]{inputenc}
\usepackage{lmodern}
\usepackage[margin=1in]{geometry}
\usepackage{amsmath,amssymb,amsfonts}
\usepackage{graphicx}
\usepackage{booktabs}
\usepackage{multirow}
\usepackage{array}
\usepackage{makecell}
\usepackage{subcaption}
\usepackage{algorithm}
\usepackage{algpseudocode}
\usepackage{url}
\usepackage{enumitem}
\usepackage{tabularx}
\usepackage[table]{xcolor}
\usepackage{adjustbox}   
\usepackage{authblk}     
\usepackage{natbib}
\usepackage{microtype}
\usepackage[colorlinks=true,linkcolor=blue,citecolor=blue,urlcolor=blue]{hyperref}

\setkeys{Gin}{keepaspectratio}

\definecolor{deepred}{RGB}{170,0,0}
\definecolor{deepblue}{RGB}{0,0,170}
\definecolor{deepgreen}{RGB}{0,120,0}
\newcommand{\rankone}[1]{\textcolor{deepred}{\textbf{#1}}}
\newcommand{\ranktwo}[1]{\textcolor{deepblue}{\textbf{#1}}}
\newcommand{\rankthree}[1]{\textcolor{deepgreen}{\textbf{#1}}}
\newcommand{\first}[1]{\rankone{#1}}
\newcommand{\second}[1]{\ranktwo{#1}}
\newcommand{\third}[1]{\rankthree{#1}}

\newcommand{\glare}{\textsc{Glare}}
\newcommand{\glareu}{\textsc{Glare}-U}
\newcommand{\R}{\mathbb{R}}

\title{Learning the Graph and the Embedding Together:\\
Classifier-Independent Rewiring for Heterophilic Node Classification}

\author{Harshit Kumar}
\author{Sujan Chakraborty}
\author{Priyanka Saha}
\author{Pritam Kar}
\author{Saptarshi Bej\thanks{Corresponding author: \texttt{sbej7042@iisertvm.ac.in}}}
\affil{School of Data Science, Indian Institute of Science Education and
Research Thiruvananthapuram, Thiruvananthapuram 695551, India}

\date{}

\begin{document}

\maketitle

\begin{abstract}
Graph neural networks lose much of their advantage on heterophilic graphs,
where connected nodes often carry different labels. Graph rewiring is a
popular remedy, but rewiring methods are usually evaluated with a single
classifier, which makes it hard to tell whether the gains come from the new
topology or from that particular pairing. We propose an affinity-guided
rewiring method that estimates the graph and the node representation
together. It alternates, in the spirit of expectation maximisation, between
training a lightweight graph neural network on the current graph and
re-weighting candidate edges under a modularity objective with a
pseudo-label homophily term. Candidate edges come from a compact pool scored
by a contrastively learned node similarity and a neighbourhood-distribution
affinity. The method returns two classifier-independent outputs: a rewired
graph and a node embedding learned on it. Across six heterophilic benchmarks
and five downstream classifiers, it improves accuracy over the original graph
with normalised features in 23 of 30 classifier-dataset combinations, with a
mean gain of 5.8 points, and reduces the accuracy spread between classifiers
about fourfold. A controlled ablation shows that the two outputs are each
useful and play complementary roles: the embedding contributes most of the
accuracy gain, while the rewired graph makes different classifiers agree. A
fully unsupervised variant, which uses no labels during rewiring, retains most
of the improvement. The rewired graphs are also more homophilic and improve
label propagation and community detection.
\end{abstract}


\section{Introduction}\label{sec:intro}

Graph neural networks (GNNs) learn node representations by passing messages
along the edges of a graph. Most message-passing architectures, such as the
graph convolutional network (GCN) \citep{kipf2017gcn} and GraphSAGE
\citep{hamilton2017graphsage}, compute a node's new representation by
averaging or summing the representations of its neighbours. This is a good
inductive bias when neighbours tend to share a label, a property called
\emph{homophily}, because averaging then removes noise while keeping the class
signal. Many real graphs, however, are \emph{heterophilic}: connected nodes
often belong to different classes. Examples include word co-occurrence graphs
in which nouns connect to adjectives, rating networks in which products of
different quality are bought together, and crowd-sourcing graphs in which
reliable and unreliable workers share tasks. On such graphs the same averaging
mixes signals from different classes and can hurt more than it helps
\citep{zhu2020h2gcn, platonov2023critical}. In several benchmarks a plain
multilayer perceptron that ignores the graph is competitive with, or better
than, standard GNNs.

Two broad responses to heterophily have emerged. The first redesigns the
architecture so that it can separate information coming from dissimilar
neighbours, for example by keeping ego and neighbour representations apart,
by aggregating over several hop distances, or by learning propagation weights
that may be negative \citep{zhu2020h2gcn, abuelhaija2019mixhop, bo2021fagcn,
chien2021gprgnn, lim2021linkx, luan2022acmgcn, li2022glognn}. The second
leaves the classifier unchanged and instead modifies the graph, adding edges
that are likely to be informative and removing edges that are not. This
second approach is called \emph{graph rewiring}
\citep{topping2022curvature, karhadkar2023fosr, bi2022dhgr,
rubiomadrigal2025comfy}.

Rewiring is attractive because it is, in principle, independent of the model:
a better graph should help many classifiers, not only one. It also separates
two concerns that architecture design mixes together, namely deciding which
nodes should exchange information, and deciding how the exchanged
information is transformed. In practice, however, rewiring methods are almost
always developed and reported together with one specific classifier from their
own paper, most often a two-layer GCN. It is then difficult to separate the
contribution of the new topology from the contribution of that particular
pairing, and it remains unclear whether a method that helps a GCN will also
help an architecture that was already built for heterophily.

In this article we take the model-independent promise of rewiring seriously.
We study the following questions.

\begin{enumerate}[label=\textbf{RQ\arabic*}, leftmargin=*]
\item Can a single rewiring procedure improve a whole family of classifiers,
      including structure-aware GNNs and structure-independent models,
      roughly uniformly on heterophilic graphs?

\item Does such a procedure make the choice of the downstream classifier
      matter less?

\item If the procedure produces both a new graph and a new node
      representation, what does each of them contribute?

\item How much of the benefit depends on labels?

\item Does the rewired graph improve structural homophily, and do these
      structural changes translate into better label propagation and
      community detection?
\end{enumerate}

Our answer is \glare\ (Graph Learning through Affinity-guided REwiring). The
central idea is that the graph and the node representation should be
estimated \emph{together}. \glare\ alternates between two steps, in the spirit
of an expectation-maximisation procedure. In the first step it learns node
embeddings and soft pseudo-labels on the current graph. In the second step it
re-estimates the edge weights from those embeddings under an objective that
combines modularity with a pseudo-label homophily term. The edges are
therefore always scored against embeddings learned on the current graph, and
the embeddings are always learned on the latest graph. Because both are kept
until the end, \glare\ returns two outputs: a rewired graph and a node
embedding learned on it. Neither output depends on the downstream classifier,
and any classifier can use one or both of them.

This design differs from existing work in several ways. Most rewiring methods
for heterophily compute the new graph in one pass from fixed features or
labels and return only a graph. Graph structure learning methods do update the
graph and the embedding together, but inside one end-to-end classifier, so the
learned graph is tied to that model. A few recent methods also refine node
features together with the graph, for example by spectral denoising
\citep{linkerhagner2025jdr} or by fusing class embeddings with the input
features \citep{bose2025labelguided}. \glare\ instead learns a new embedding by
message passing on the graph that it is building, uses an edge objective
designed for heterophily, runs with or without labels, and is evaluated with
five classifiers so that the roles of its two outputs can be studied
separately.

\paragraph{Contributions.} The main contributions of this article are as
follows.
\begin{itemize}
\item We propose \glare, an affinity-guided method that alternates between
      embedding learning and edge re-estimation. Its objective couples a
      modularity term on learned edge weights with a homophily term driven by
      pseudo-labels, and it is regularised by a degree budget, an entropy
      term and a prior that anchors the original edges
      (Section~\ref{sec:method}).
\item We evaluate every configuration with five downstream classifiers of
      different design and against five rewiring and structure-learning
      baselines, each run with the classifier recommended in its own paper.
      Against the original graph with $\ell_2$-normalised features, \glare\
      improves accuracy in 23 of 30 classifier-dataset combinations, with a
      mean gain of $5.8$ points, and reduces the spread of accuracy across
      classifiers about fourfold (Sections~\ref{sec:res_main}
      and~\ref{sec:res_uniform}).
\item With a controlled $2\times2$ ablation we show that both outputs improve
      over the original graph on their own and that they do different jobs:
      the embedding gives the larger accuracy gain, while the rewired graph
      does more to make different classifiers agree
      (Section~\ref{sec:res_ablation}).
\item We introduce \glareu, a fully unsupervised variant that uses no label of
      any split during rewiring, and show that it keeps most of the gain
      (Section~\ref{sec:res_unsup}).
\item We analyse the rewired graphs beyond classification accuracy, with five
      homophily measures, label propagation, community detection and linear
      probing of the embeddings (Section~\ref{sec:res_structure}), and we
      discuss in detail when \glare\ does not help
      (Section~\ref{sec:discussion}).
\end{itemize}

The rest of the article is organised as follows. Section~\ref{sec:background}
introduces notation and the homophily measures we use, and reviews related
work on heterophily-aware architectures, graph rewiring, graph structure
learning and joint refinement of graphs and features.
Section~\ref{sec:method} describes \glare. Section~\ref{sec:setup} gives the
experimental setup and Section~\ref{sec:results} the results.
Section~\ref{sec:discussion} discusses the findings, practical guidance and
limitations, and Section~\ref{sec:conclusion} concludes.

\section{Background and related work}\label{sec:background}

\subsection{Notation and problem setting}\label{sec:notation}
Let $\mathcal{G}=(X,A)$ be an undirected graph with $N$ nodes, node features
$X\in\R^{N\times F}$ and adjacency matrix $A\in\{0,1\}^{N\times N}$. We write
$\mathcal{E}$ for the edge set, $d_u$ for the degree of node $u$ and
$\mathcal{N}(u)$ for its neighbours. Each node has a label
$y_u\in\{1,\dots,C\}$, observed only on a training set
$\mathcal{V}_{\text{tr}}$; validation and test sets are disjoint from it. We
write $Y_L$ for the visible training labels. In semi-supervised node
classification the goal is to predict the labels of the test nodes. A
\emph{rewiring} method maps $(X,A,Y_L)$ to a new adjacency $\tilde A$, which is
then given to a downstream classifier. In this article a rewiring method may
also return a new node representation $Z\in\R^{N\times D}$.

\subsection{Measuring homophily}\label{sec:homophily_measures}
Several measures of homophily are in use, and they can disagree, so we report
five of them. \emph{Edge homophily} is the fraction of edges that connect
nodes of the same class \citep{zhu2020h2gcn},
\begin{equation}
h_{\text{edge}}=\frac{|\{(u,v)\in\mathcal{E}: y_u=y_v\}|}{|\mathcal{E}|}.
\end{equation}
\emph{Node homophily} averages the same fraction per node
\citep{pei2020geomgcn}, $h_{\text{node}}=\frac{1}{N}\sum_u
|\{v\in\mathcal{N}(u):y_v=y_u\}|/d_u$. Both depend on the class balance: a
graph with one dominant class has a high edge homophily even if edges are
random. \emph{Class-insensitive edge homophily} \citep{lim2021linkx} corrects
for this by comparing, for each class $k$, the fraction $h_k$ of same-class
edges leaving class-$k$ nodes with the class proportion,
\begin{equation}
h_{\text{CI}}=\frac{1}{C-1}\sum_{k=1}^{C}\max\!\Big(0,\;h_k-\frac{|\mathcal{V}_k|}{N}\Big).
\end{equation}
\emph{Adjusted homophily} \citep{platonov2023characterizing} subtracts the
value expected under a degree-preserving random graph,
\begin{equation}
h_{\text{adj}}=\frac{h_{\text{edge}}-\sum_k \bar p_k^{\,2}}{1-\sum_k \bar p_k^{\,2}},
\qquad \bar p_k=\frac{\sum_{u:y_u=k} d_u}{2|\mathcal{E}|},
\end{equation}
and can be negative. Finally, \emph{label informativeness}
\citep{platonov2023characterizing} measures how much the label of a node
tells about the label of a random neighbour,
$\mathrm{LI}=I(y_\xi;y_\eta)/H(y_\xi)$ for the endpoints $(\xi,\eta)$ of a
random edge. A graph can be heterophilic yet informative, for example when
every class connects to one specific other class. We also report a feature
homophily score $h_{\text{feat}}$, the mean cosine similarity of the raw
features of connected nodes, which shows how well the features align with the
edges.

\subsection{GNN architectures for heterophily}\label{sec:rw_arch}
Early work on heterophily changed the aggregation itself. Geom-GCN
\citep{pei2020geomgcn} aggregates over neighbourhoods defined in a latent
geometric space. MixHop \citep{abuelhaija2019mixhop} mixes representations
from several hop distances, and H2GCN \citep{zhu2020h2gcn} combines three
design choices: separating ego and neighbour embeddings, using higher-order
neighbourhoods, and combining intermediate representations. FAGCN
\citep{bo2021fagcn} learns signed attention coefficients so that it can use
both low- and high-frequency signals, and GPR-GNN \citep{chien2021gprgnn}
learns generalised PageRank weights that may be negative. LINKX
\citep{lim2021linkx} embeds the adjacency and the features separately with
multilayer perceptrons and combines them, which scales well and is strong on
large heterophilic graphs. ACM-GCN \citep{luan2022acmgcn} adaptively mixes
low-pass, high-pass and identity channels, and GloGNN \citep{li2022glognn}
aggregates over all nodes with learned coefficients. \citet{zhu2021heterophily_survey}
give a broader overview. \citet{platonov2023critical} showed that several of
the earlier heterophilic benchmarks contain duplicated nodes that leak
information between training and test sets, released new heterophilic
datasets, and found that standard GNNs with residual connections are strong
baselines once evaluation is done carefully. These architectures and
\glare\ are complementary: two of our downstream classifiers, H2GCN and LINKX,
are heterophily-aware, and one question of this article is whether rewiring
still helps them.

\subsection{Graph rewiring}\label{sec:rw_rewiring}
Graph rewiring was first studied for problems other than heterophily.
\emph{Over-smoothing} makes node representations indistinguishable as depth
grows \citep{li2018deeper}, and \emph{over-squashing} compresses information
from exponentially growing neighbourhoods into fixed-size vectors
\citep{alon2021bottleneck}. Curvature-based methods such as SDRF
\citep{topping2022curvature} and BORF \citep{nguyen2023borf} add edges around
negatively curved bottlenecks and remove edges in positively curved regions.
Spectral methods such as FoSR \citep{karhadkar2023fosr} add edges that
increase the spectral gap, and DiffWire \citep{arnaizrodriguez2022diffwire}
learns rewiring layers based on commute times and the spectral gap. Diffusion-based
approaches such as graph diffusion convolution \citep{gasteiger2019diffusion}
and APPNP \citep{gasteiger2019appnp} replace the adjacency matrix by a sparsified
personalised PageRank matrix, which smooths the graph and removes noisy
edges. These methods do not use labels and do not target heterophily
directly.

Rewiring for heterophily tries instead to increase homophily. DHGR
\citep{bi2022dhgr} measures the similarity of the label and feature
distributions in node neighbourhoods and uses it to add homophilic edges and
prune heterophilic ones. LPkG \citep{park2024lpkg} builds a $k$-nearest
neighbour graph and propagates labels on it to select edges. ComFy
\citep{rubiomadrigal2025comfy} rewires using community structure together with
feature similarity, and label-guided rewiring \citep{bose2025labelguided}
fuses class embeddings from a dense network with the node features before
rewiring. Most of these methods compute the new graph once, before training,
from fixed features or labels, return only a graph, and are evaluated with
one classifier.

\subsection{Graph structure learning}\label{sec:rw_gsl}
Graph structure learning treats the adjacency matrix as a parameter to be
learned. LDS \citep{franceschi2019lds} learns a discrete edge distribution by
bilevel optimisation, Pro-GNN \citep{jin2020prognn} learns a clean graph that
is sparse, low-rank and feature-smooth in order to resist adversarial attacks,
and IDGL \citep{chen2020idgl} iteratively refines a similarity graph and the
node embeddings until they are consistent. SLAPS \citep{fatemi2021slaps} adds a
self-supervised denoising task that provides extra supervision for the
learned adjacency. These methods alternate between graph and embedding, as
\glare\ does, but they optimise both together with one classifier, and the
learned graph mostly connects nodes whose features or embeddings are similar,
which is itself a homophily assumption. SUBLIME \citep{liu2022sublime} learns
the structure without labels by bootstrapped contrastive learning between a
learned view and an anchor view of the graph. Contrastive objectives on graphs
\citep{zhu2020grace} are also the basis of our similarity encoder.

\subsection{Joint refinement of graph and features}\label{sec:rw_joint}
The closest work to ours refines both the graph and the node features before
classification. JDR \citep{linkerhagner2025jdr} assumes that the graph and the
features are noisy views of the same latent class structure. It rewires the
graph and denoises the features by iteratively aligning their leading
spectral subspaces, a strategy motivated by the contextual stochastic block
model, and it reports that both parts help downstream GNNs. Label-guided
rewiring \citep{bose2025labelguided} also passes enriched features together
with the rewired graph. \glare\ shares with these methods the idea of
returning two outputs, but its second output is not a cleaned copy of the
input features: it is a new representation learned by message passing on the
graph that the loop is building, so that the graph and the embedding are
refined against each other. \glare\ also differs in its objective, which is
written for heterophily, in having a fully unsupervised variant, and in our
evaluation, which studies the two outputs separately across five
classifiers.

\subsection{Positioning}\label{sec:positioning}
Table~\ref{tab:positioning} summarises how \glare\ relates to the main lines
of work. \glare\ keeps the alternating graph and embedding updates of
structure learning, but runs them as a stand-alone stage in front of any
classifier, as rewiring does. Its edge objective looks for edges between nodes
of the same class, and not only between nodes with similar features, and it
gives two reusable outputs instead of a single graph or a single trained model.

\begin{table}[htbp]
\centering
\caption{Positioning of \glare\ with respect to representative methods.
``Coupled'' means that the graph is re-estimated against representations that
are themselves updated, and ``Joint with classifier'' means that the learned
graph is optimised inside one end-to-end classifier.}
\label{tab:positioning}
\begin{adjustbox}{max width=\textwidth}
\begin{tabular}{@{}llccc@{}}
\toprule
Method & Primary goal & Output & Coupled & Joint with classifier \\
\midrule
SDRF, BORF, FoSR & over-squashing & graph & no & no \\
DHGR, LPkG, ComFy & heterophily & graph & no & no \\
IDGL, SLAPS & structure learning & graph (tied to model) & yes & yes \\
\glare\ (ours) & heterophily, any classifier & graph + embedding & yes & no \\
\bottomrule
\end{tabular}
\end{adjustbox}
\end{table}

\section{The \glare\ method}\label{sec:method}
\begin{figure}[!t]
    \centering
    \includegraphics[width=\textwidth,height=0.8\textheight]{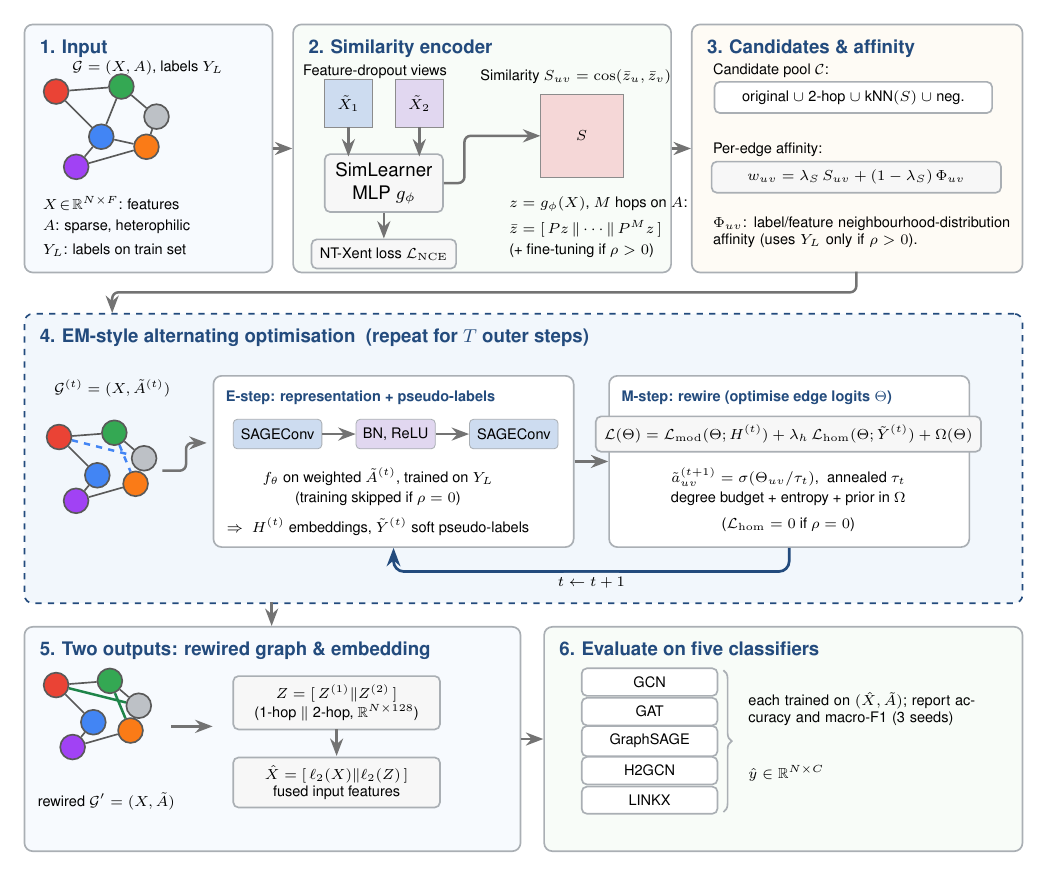}
    \caption{\glare\ pipeline. A contrastively pretrained encoder produces a
    node similarity $S$; a compact candidate pool is scored by an affinity that
    blends $S$ with a neighbourhood-distribution signal $\Phi$; an EM-style loop
    then alternates between training a lightweight GNN on the current soft graph
    (E-step) and updating per-edge logits under a modularity-with-homophily
    objective (M-step). The rewired graph and a multi-hop representation are
    finally passed to any downstream classifier.}
    \label{fig:model}
\end{figure}

\glare\ takes a graph $\mathcal{G}=(X,A)$ and, optionally, visible training
labels $Y_L$, and returns a rewired adjacency $\tilde A$ and a node embedding
$Z$. The amount of label information is controlled by a mask ratio
$\rho\in[0,1]$: $\rho=1$ uses all training labels and $\rho=0$ uses none.
Figure~\ref{fig:model} shows the pipeline and Algorithm~\ref{alg:glare} gives
the pseudo-code. The method has four stages: a similarity encoder
(Section~\ref{sec:encoder}), a candidate pool with an affinity score
(Section~\ref{sec:candidates}), an alternating optimisation of the edge
weights (Section~\ref{sec:em}), and the construction of the two outputs
(Section~\ref{sec:fusion}). We end the section with the unsupervised variant
(Section~\ref{sec:glareu}) and a complexity analysis
(Section~\ref{sec:complexity}).

\subsection{Similarity encoder}\label{sec:encoder}
Rewiring needs a notion of which pairs of nodes are likely to share a label.
Raw features are often a poor guide: on heterophilic graphs they can be
sparse, high-dimensional and weakly aligned with the classes (the feature
homophily $h_{\text{feat}}$ in Table~\ref{tab:dataset_stats} is below $0.02$
for Chameleon and Squirrel). We therefore first learn a node embedding
$z_u=g_\phi(X_u)$ with an encoder $g_\phi$ made of blocks of linear,
BatchNorm and PReLU layers with an input-to-output residual connection. Two
feature-dropout views $\tilde X_1,\tilde X_2$ of the features are encoded, and
$g_\phi$ is trained with the NT-Xent contrastive loss \citep{chen2020simclr},
which pulls the two views of the same node together and pushes the views of
different nodes apart. This pretraining uses neither labels nor edges.

The embeddings are then propagated $M$ hops over the input graph with the
row-normalised adjacency $P$ and the hops are concatenated,
$\bar z=[\,Pz\,\Vert\,\dots\,\Vert\,P^{M}z\,]$. The similarity of two nodes is
the cosine of their mean-centred propagated embeddings,
$S_{uv}=\cos(\bar z_u,\bar z_v)$. Propagation makes the similarity reflect the
neighbourhood of a node and not only its own features, which is important on
heterophilic graphs, where two nodes of the same class often have similar
neighbourhoods even when they are not connected. When training labels are
visible ($\rho>0$), $g_\phi$ is additionally fine-tuned so that the similarity
of labelled pairs under $\bar z$ matches the similarity of their $M$-hop
propagated label distributions. With $\rho=0$ this fine-tuning is skipped
and $S$ uses no labels.

\subsection{Candidate edges and affinity}\label{sec:candidates}
Considering all $N^2$ node pairs is infeasible for large graphs and also not
needed, since most pairs are clearly unrelated. We form a compact candidate
pool
\begin{equation}
\mathcal{C}=\mathcal{E}\;\cup\;\mathcal{E}_{\text{2-hop}}\;\cup\;
\mathcal{E}_{\text{feat}}\;\cup\;\mathcal{E}_{k\text{NN}(S)}\;\cup\;\mathcal{E}_{\text{rand}},
\end{equation}
which contains the original edges; up to $B/2$ two-hop pairs, where $B$ is a
candidate budget ($B=10^4$ in all experiments); pairs of nodes with similar raw
features, namely the five nearest neighbours of a set of probe nodes (all
nodes for graphs with at most $3000$ nodes, otherwise a degree-weighted sample
of $\max(512,\sqrt{N})$ nodes) and similar non-adjacent pairs that fill the
rest of the budget; the $k$ nearest neighbours of every node under the learned
similarity $S$ with $S_{uv}$ above a small threshold; and, only if the pool is
still small, a few random pairs. Each part has a role. Original edges let
\glare\ keep what is already useful. Two-hop pairs are a natural source of
same-class edges on heterophilic graphs, because two neighbours of a node
often share a class even when the node itself does not. Nearest neighbours
under the raw features and under $S$ connect nodes that are similar but far
apart in the graph, and random pairs give the objective negative examples.

Each candidate $(u,v)\in\mathcal{C}$ receives an affinity
\begin{equation}
  w_{uv}=\lambda_S\, S_{uv}+(1-\lambda_S)\,\Phi_{uv},
  \label{eq:affinity}
\end{equation}
where $\Phi_{uv}$ is a neighbourhood-distribution affinity. For visible labels
we propagate a one-hot label matrix $M$ hops, and we do the same with the
features; we then compare the resulting distributions of $u$ and $v$ with a
de-centred cosine. A coverage-based confidence blends the label signal, where
propagated labels reach, with the feature signal elsewhere. Pairs of two
visible nodes of the same class are given a hard positive anchor. Visible
training labels therefore enter the rewiring in four places: $\Phi$ and its
anchors, the fine-tuning of $g_\phi$ (Section~\ref{sec:encoder}), the training
of the E-step network, and the trust weights of the homophily term
(Section~\ref{sec:em}). With $\rho=0$ all four are inactive.

\subsection{Alternating optimisation of the edge weights}\label{sec:em}
We attach to every candidate a logit $\Theta_{uv}$ and define a soft edge
weight $\tilde a_{uv}=\sigma(\Theta_{uv}/\tau)$ with temperature $\tau$. The
logits are initialised from a prior $\sigma(\kappa\,w_{uv})$ that also floors
the weights of the original edges, so \glare\ starts close to the input graph
and departs from it only when the objective rewards it. \glare\ then
alternates, for $T$ outer iterations, between an E-step and an M-step.

\paragraph{E-step: representation and pseudo-labels.}
Given the current soft graph $\mathcal{G}^{(t)}=(X,\tilde A^{(t)})$, we train a
two-layer weighted GraphSAGE network $f_\theta$, a self-and-neighbour
aggregation that accepts continuous edge weights, on the visible training
labels with gradient clipping. From $f_\theta$ we read node embeddings
$H^{(t)}$ and soft pseudo-labels
$\tilde Y^{(t)}=\mathrm{softmax}(f_\theta(X,\tilde A^{(t)}))$. The network is
warm-started from the previous outer iteration, so that the representation
changes smoothly as the graph changes. When no training labels are visible
($\rho=0$), $f_\theta$ is not trained, and $H^{(t)}$ and $\tilde Y^{(t)}$ come
from its randomly initialised weights.

\paragraph{M-step: rewiring.}
Holding $H^{(t)}$ and $\tilde Y^{(t)}$ fixed, we update the edge logits
$\Theta$ by minimising
\begin{equation}
  \mathcal{L}(\Theta)=\underbrace{\mathcal{L}_{\mathrm{mod}}(\Theta;H^{(t)})}_{\text{structure}}
  +\;\lambda_h\,\underbrace{\mathcal{L}_{\mathrm{hom}}(\Theta;\tilde Y^{(t)})}_{\text{pseudo-label homophily}}
  +\;\Omega(\Theta).
  \label{eq:objective}
\end{equation}
The structural term is a negative modularity functional
\citep{newman2006modularity}. Classical modularity rewards edges inside
communities relative to a degree-preserving null model. Here the ``community''
signal is replaced by an embedding-similarity gain
$g_{uv}=\beta\cos(H_u,H_v)+(1-\beta)\Phi_{uv}$, so that the term rewards
weight on candidate edges whose endpoints are more similar than the null model
predicts from their soft degrees. The homophily term rewards weight on
candidate edges whose endpoints share a pseudo-label, weighted by a trust
score, which is the product of the endpoints' label coverage, or $1$ for
anchored pairs. It therefore vanishes when $\rho=0$. The regulariser
\begin{equation}
\Omega=\lambda_e\Omega_{\text{ent}}+\lambda_p\Omega_{\text{prior}}+\lambda_d\Omega_{\text{deg}}
\end{equation}
collects an entropy term, which drives the weights towards $0$ or $1$, a prior
term, which keeps the weights close to the affinity $w_{uv}$, and a
degree-budget term, which controls the density of the rewired graph and
prevents trivial solutions that keep all or no edges. The temperature $\tau$
is annealed with a cosine schedule, so that early iterations explore soft
graphs and late iterations commit to a nearly binary topology. Only $\Theta$
is updated in the M-step and only $\theta$ in the E-step.

\paragraph{Model selection.}
For \glare\ we evaluate $f_\theta$ on the thresholded graph after each outer
iteration and keep the pair $(\theta,\Theta)$ with the best validation
accuracy. \glareu\ keeps the last iterate, so it never reads a validation
label (Section~\ref{sec:glareu}).

\paragraph{Why alternate?}
The two steps solve two coupled problems. A good graph needs a representation
that tells which nodes should be connected, and a good representation needs a
graph on which message passing does not mix classes. Solving either problem
with the other held fixed at its initial value, as one-pass rewiring does with
the input features, ignores this coupling. Alternating lets each estimate
improve the other: edges between nodes that the current representation
considers similar are strengthened, which makes the next representation more
class-consistent, which in turn gives a better similarity for the next
rewiring step. The ablation in Section~\ref{sec:res_ablation} and the
unsupervised variant in Section~\ref{sec:res_unsup} examine how much each part
contributes in practice.

\subsection{Two outputs: rewired graph and embedding}\label{sec:fusion}
After the loop we keep the candidate edges whose weight exceeds a threshold
and, if the result is denser than a target of $0.7$ times the number of
original edges, we keep the edges whose endpoints have the most similar
embeddings until the target is met. This gives the rewired graph $\tilde A$.
We then compute a two-hop representation
$Z=[Z^{(1)}\Vert Z^{(2)}]\in\R^{N\times 128}$ on $\tilde A$, where $Z^{(1)}$ is
the first hidden layer of $f_\theta$ and $Z^{(2)}$ is one more
degree-normalised propagation of $Z^{(1)}$ followed by a ReLU activation.
Since $Z$ is computed on $\tilde A$, it carries the structure found by the
loop, and it remains useful even when message passing uses the original edges
(Section~\ref{sec:res_ablation}). We $\ell_2$-normalise and concatenate the
original features and this representation, $\hat X=[\ell_2(X)\Vert\ell_2(Z)]$,
and feed $(\hat X,\tilde A)$ to the downstream classifier. The rewired graph
and the embedding are the only things passed on; no part of the classifier is
built into \glare, and a user can take only one of the two outputs.

\subsection{Fully unsupervised variant}\label{sec:glareu}
Setting $\rho=1$ gives \glare, the main method. Setting $\rho=0$ and keeping the
last iterate gives \glareu. In \glareu\ no label of any split reaches the
rewiring: $\Phi$ reduces to its feature signal, the encoder is not fine-tuned,
the E-step network keeps its random initial weights and the homophily term
vanishes. The graph is therefore built from the contrastive encoder, the
feature-based affinity and the modularity objective only, and $Z$ is a
random-weight propagation of the features over the rewired graph. Only the
downstream classifiers use training labels, exactly as for every other method
in our comparison. \glareu\ is useful in two ways. In practice it is a rewiring
method that can be run before any labels are available, and scientifically it
isolates the contribution of the label-free parts of the method.

\begin{algorithm}[htbp!]
\caption{\glare\ rewiring}
\label{alg:glare}
\begin{algorithmic}[1]
\Require graph $\mathcal{G}=(X,A)$, visible labels $Y_L$, mask ratio $\rho$,
         outer iterations $T$
\State train encoder $g_\phi$ with NT-Xent on feature-dropout views;
       if $\rho>0$, fine-tune on visible labels; set
       $S_{uv}=\cos(\bar z_u,\bar z_v)$ with $\bar z$ the $M$-hop propagated
       embeddings
\State build candidate pool $\mathcal{C}$; compute $\Phi$ from $\rho$-masked
       label/feature propagation; set $w_{uv}$ by Eq.~\eqref{eq:affinity}
\State initialise logits $\Theta \gets \mathrm{logit}(\sigma(\kappa\,w))$,
       floor original edges
\For{$t=0$ to $T-1$}
  \State anneal temperature $\tau_t$ (cosine)
  \State \textbf{E-step:} if $\rho>0$, train weighted GNN $f_\theta$ on
         $\mathcal{G}^{(t)}=(X,\sigma(\Theta/\tau_t))$ with visible labels;
         read $H^{(t)}$, $\tilde Y^{(t)}$
  \State \textbf{M-step:} update $\Theta$ by minimising
         Eq.~\eqref{eq:objective}
  \State \glare: keep $(\theta,\Theta)$ with best validation accuracy so
         far; \glareu: keep the current $(\theta,\Theta)$
\EndFor
\State threshold weights to get $\tilde A$; compute $Z=[Z^{(1)}\Vert Z^{(2)}]$
       on $\tilde A$; set $\hat X=[\ell_2(X)\Vert\ell_2(Z)]$
\State \Return rewired graph $\tilde A$ and fused features $\hat X$
\end{algorithmic}
\end{algorithm}

\subsection{Computational complexity}\label{sec:complexity}
Let $D$ be the hidden size and $|\mathcal{C}|$ the size of the candidate pool.
Apart from the $|\mathcal{E}|$ original edges and the $kN$ pairs from the
learned similarity, the pool is bounded by the fixed budget $B$ and the probe
set, so $|\mathcal{C}|=O(|\mathcal{E}|+kN+B)$. The
encoder costs $O(e_s N F D)$ for $e_s$ pretraining epochs, and the
nearest-neighbour search computes similarities between all pairs of nodes in
blocks of $b$ rows, which costs $O(N^2 D)$ time but only $O(bN)$ memory at a time. Each
outer iteration trains $f_\theta$ for $e$ epochs on the weighted candidate
graph, at cost $O(e(|\mathcal{C}|D+NFD))$, and runs $r$ gradient steps on
$\Theta$, at cost $O(r|\mathcal{C}|)$ once the gains $g_{uv}$ are computed in
$O(|\mathcal{C}|D)$. The total cost of the loop is therefore
$O\big(T[(e+r)|\mathcal{C}|D+eNFD]\big)$, linear in the number of candidate
edges, and the memory is $O(|\mathcal{C}|+N(F+D))$. In our experiments the
nearest-neighbour step and the $T\cdot e$ epochs of the E-step dominate the
running time; \glareu\ skips the E-step training and is correspondingly much
faster (Section~\ref{sec:res_unsup}).

\section{Experimental setup}\label{sec:setup}

\subsection{Datasets}\label{sec:datasets}
We use six heterophilic node-classification benchmarks
(Table~\ref{tab:dataset_stats}). \emph{Actor} \citep{pei2020geomgcn} is the
actor-only subgraph of a film-director-actor-writer network, with edges
between actors who appear on the same Wikipedia page. \emph{Chameleon} and
\emph{Squirrel} \citep{rozemberczki2021musae} are networks of Wikipedia pages
connected by mutual links, with classes given by average monthly traffic.
\emph{Roman-empire}, \emph{Amazon-ratings} and \emph{Tolokers}
\citep{platonov2023critical} are, respectively, a word graph built from the
Wikipedia article on the Roman Empire with syntactic roles as classes, a
co-purchase graph with product ratings as classes, and a crowd-sourcing graph
of workers who shared tasks, with the task of predicting banned workers.
Tolokers is binary and imbalanced: $78.2\%$ of the test nodes belong to the
majority class.

The datasets span very different sizes (2{,}277 to 24{,}492 nodes), densities
(mean degree 2.9 to 88.3) and numbers of classes (2 to 18). All six are
strongly heterophilic under every structural measure in
Table~\ref{tab:dataset_stats}, and they also differ in how informative the
features are: the feature homophily is high on Tolokers and Amazon-ratings and
close to zero on Chameleon and Squirrel. For Chameleon and Squirrel we use the
original graphs, which \citet{platonov2023critical} showed to contain
duplicated nodes that can leak information between the training and test
sets; results on these two datasets should therefore be read with this caveat
in mind.

\begin{table}[htbp!]
\centering
\caption{%
  Properties of the six heterophilic benchmark datasets with their original homophily scores.
  $|V|$, $|E|$, $C$ and $\bar{d}$ denote node count, directed edge count,
  number of class labels, and mean node degree, respectively.
  All structural homophily metrics confirm strong heterophily
  ($h_{\text{edge}}, h_{\text{node}}, h_{\text{adj}}, h_{\text{CI}} \ll 1$).
  $h_{\text{feat}}$ is the feature homophily computed on original node features $\mathbf{X}$.
  Datasets are ordered from least to most structurally homophilic (by $h_{\text{edge}}$).
}
\label{tab:dataset_stats}
\setlength{\tabcolsep}{5pt}
\begin{adjustbox}{max width=\textwidth}
\begin{tabular}{l r r r r r r r r r r}
\toprule
\textbf{Dataset}
  & $|V|$ & $|E|$ & $C$ & $\bar{d}$
  & $h_{\text{edge}}$ & $h_{\text{node}}$
  & $h_{\text{adj}}$  & $h_{\text{CI}}$
  & LI     & $h_{\text{feat}}$ \\
\midrule
Roman-empire    & 22{,}662 &   65{,}854 & 18 &  2.91 & 0.047 & 0.046 & $-$0.047 & 0.021 & 0.110 & 0.213 \\
Actor           &  7{,}600 &   30{,}019 &  5 &  3.95 & 0.219 & 0.225 &    0.006 & 0.012 & 0.000 & 0.162 \\
Squirrel      &  5{,}201 &  217{,}073 &  5 & 41.74 & 0.224 & 0.210 &    0.012 & 0.026 & 0.002 & 0.016 \\
Chameleon     &  2{,}277 &   36{,}101 &  5 & 15.85 & 0.235 & 0.274 &    0.039 & 0.063 & 0.052 & 0.017 \\
Amazon-ratings  & 24{,}492 &  186{,}100 &  5 &  7.60 & 0.380 & 0.376 &    0.140 & 0.127 & 0.040 & 0.522 \\
Tolokers        & 11{,}758 & 1{,}038{,}000 & 2 & 88.28 & 0.595 & 0.634 & 0.093 & 0.180 & 0.007 & 0.809 \\
\bottomrule
\end{tabular}
\end{adjustbox}
\end{table}

\subsection{Downstream classifiers}\label{sec:classifiers}
We evaluate every graph configuration with five classifiers that cover
different design choices: GCN \citep{kipf2017gcn}, which averages normalised
neighbour representations; GAT \citep{velickovic2018gat}, which learns
attention weights over neighbours; GraphSAGE \citep{hamilton2017graphsage},
which keeps a separate self representation; H2GCN \citep{zhu2020h2gcn}, which
separates ego and neighbour embeddings and uses two-hop neighbourhoods; and
LINKX \citep{lim2021linkx}, which embeds the adjacency and the features
separately. The first three assume homophily and the last two were designed
for heterophily. Each classifier is given an input-to-logits residual path, so
that all five share the same training harness: hidden size 128, dropout 0.5,
label smoothing 0.1, Adam \citep{kingma2015adam} with cosine learning-rate
annealing, gradient clipping at 5.0, 300 epochs, and model selection on
validation accuracy every five epochs. On the original graph the classifiers
receive the $\ell_2$-normalised features $\ell_2(X)$, which is the same
normalisation used inside $\hat X$ and, as Section~\ref{sec:res_raw} shows, a
stronger reference than raw features.

\subsection{Baselines}\label{sec:baselines}
We compare with five rewiring and structure-learning methods, each run with
the classifier recommended in its own paper (Table~\ref{tab:baseclf}), since
that pairing is part of the method. \emph{DHGR} \citep{bi2022dhgr} adds and
prunes edges according to the similarity of neighbourhood label and feature
distributions. \emph{LPkG} \citep{park2024lpkg} propagates labels on a
$k$-nearest-neighbour graph built with a graph autoencoder. \emph{ComFy}
\citep{rubiomadrigal2025comfy} rewires with community structure and feature
similarity. \emph{FoSR} \citep{karhadkar2023fosr} adds edges that increase the
spectral gap and represents rewiring designed for over-squashing.
\emph{IDGL} \citep{chen2020idgl} learns the graph jointly with a GCN and
represents graph structure learning. We did not include JDR
\citep{linkerhagner2025jdr} in this study; we discuss this in
Section~\ref{sec:limitations}.

\begin{table}[htbp!]
  \centering
  \caption{Competing methods and the downstream classifier each uses, as
  recommended in its own paper. \glare\ is the only method evaluated across
  all five classifiers.}
  \label{tab:baseclf}
  \small
  \begin{tabularx}{\textwidth}{@{}lXX@{}}
    \toprule
    Method & Type & Supporting classifier \\
    \midrule
    IDGL~\citep{chen2020idgl}
      & Graph structure learning
      & 2-layer GCN (joint) \\

    FoSR~\citep{karhadkar2023fosr}
      & Spectral rewiring
      & GCN \\

    ComFy~\citep{rubiomadrigal2025comfy}
      & Community/feature rewiring
      & GCN \\

    LPkG~\citep{park2024lpkg}
      & $k$-NN + label propagation
      & GAE + LP + GNN \\

    DHGR~\citep{bi2022dhgr}
      & Heterophily rewiring
      & GCN \\

    \midrule

    \glare\ (ours)
      & Affinity-guided rewiring
      & GCN, GAT, SAGE, H2GCN, LINKX \\

    \bottomrule
  \end{tabularx}
\end{table}
\subsection{Protocol and metrics}\label{sec:protocol}
We use the first of the standard public train/validation/test splits of each
dataset (the Geom-GCN splits for Actor, Chameleon and Squirrel, and the splits
of \citet{platonov2023critical} for the other three) and report the mean and
standard deviation of test accuracy over three seeds $\{0,1,2\}$. For Tolokers
we additionally report the area under the receiver operating characteristic
curve (ROC-AUC), because accuracy is close to the majority-class rate for all
methods. Rewiring is run once per dataset and seed and cached, then reused
across the five classifiers and in the ablations. Test labels are never used
during rewiring. For supervised \glare, training labels and the validation set
are used as described in Section~\ref{sec:method}; for \glareu\ no labels are
used.

\subsection{Implementation}\label{sec:implementation}
All models are implemented in PyTorch Geometric \citep{fey2019pyg}.
Table~\ref{tab:hparams} lists the default \glare\ hyperparameters, which are
shared by all datasets. Baselines use the hyperparameters of their original
papers.

\begin{table}[htbp!]
  \centering
  \caption{Default \glare\ hyperparameters.}
  \label{tab:hparams}
  \small
  \begin{adjustbox}{max width=\textwidth}
  \begin{tabular}{lll}
    \toprule
    Group & Parameter & Value \\
    \midrule
    \multirow{3}{*}{Similarity encoder}
      & hidden size & 64 \\
      & NT-Xent temperature & 0.5 \\
      & feature-dropout rate & 0.2 \\
    \midrule
    \multirow{4}{*}{Candidates \& affinity}
      & candidate budget $B$ & 10{,}000 \\
      & $k$ (kNN on $S$) & 8 \\
      & distribution hops $M$ & 2 \\
      & learned-affinity weight $\lambda_S$ & 0.5 \\
    \midrule
    \multirow{7}{*}{EM loop}
      & outer iterations $T$ & 25 \\
      & inner GNN epochs & 80 \\
      & rewire steps / outer & 120 \\
      & GNN hidden size & 64 \\
      & homophily weight $\lambda_h$ & 0.25 \\
      & structure mix $\beta$ & 0.5 \\
      & original-edge retain floor & 0.3 \\
    \midrule
    \multirow{2}{*}{Output graph}
      & edge-weight threshold & 0.5 \\
      & target edge ratio (w.r.t.\ original) & 0.7 \\
    \midrule
    \multirow{3}{*}{Regularisation}
      & entropy weight & 0.02 \\
      & prior weight & 0.10 \\
      & degree-budget weight & 0.05 \\
    \bottomrule
  \end{tabular}
  \end{adjustbox}
\end{table}


\section{Results}\label{sec:results}
We organise the results around the five research questions of
Section~\ref{sec:intro}. Section~\ref{sec:res_main} compares \glare\ with the
baselines, Section~\ref{sec:res_uniform} studies uniformity across classifiers
(RQ1, RQ2), Section~\ref{sec:res_auc} checks the imbalanced Tolokers dataset
with ROC-AUC, Section~\ref{sec:res_ablation} separates the two outputs (RQ3),
Section~\ref{sec:res_unsup} studies the role of labels (RQ4),
Section~\ref{sec:res_structure} examines whether rewiring produces a more
class-consistent graph through structural homophily, label propagation, and
community detection (RQ5), and Section~\ref{sec:res_raw} reports the effect of
feature normalisation on the reference.

\subsection{Comparison with the baselines}\label{sec:res_main}
Table~\ref{tab:main} puts all results together: the five baselines, each with
its own classifier, the five classifiers on the original graph with
$\ell_2$-normalised features, and the same five classifiers on the \glare\
outputs. Among the rewiring methods, \glare\ gives the best accuracy on Actor,
Roman-empire and Tolokers, comes within $0.2$ points of the best baseline
(LPkG) on Amazon-ratings, and trails DHGR on Chameleon and Squirrel. Every
\glare\ configuration has a higher mean accuracy over the six datasets than
every baseline ($0.571$ to $0.574$, against at most $0.554$ for DHGR), and so
does the fully unsupervised \glareu\ ($0.562$), even though DHGR, LPkG and IDGL
use training labels to build their graphs.

\begin{figure}[htbp!]
  \centering
  \includegraphics[width=\textwidth,height=0.8\textheight]{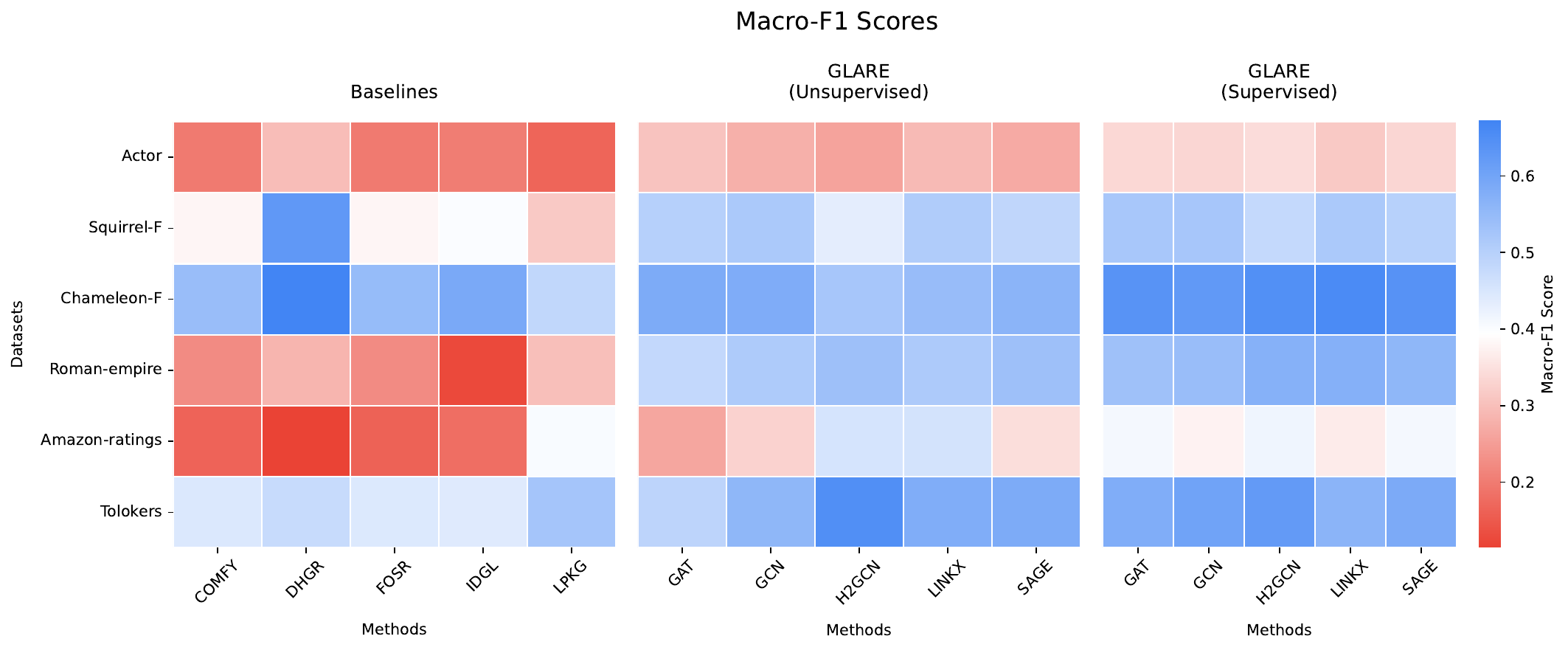}
  \caption{Macro-F1 scores of the baseline models, the fully unsupervised
  \glareu\ and \glare\ across the six datasets. The heatmap provides a complementary
  view of the baseline accuracy comparison, showing the performance of
  each rewiring method and the downstream classifiers used with \glare.}
  \label{fig:macro_f1}
\end{figure}

Two observations qualify this comparison. First, on Roman-empire and
Amazon-ratings the strongest classifiers on the original graph (GraphSAGE and
H2GCN on Roman-empire, LINKX and H2GCN on Amazon-ratings) are better than
every rewiring method, including \glare. We come back to this in
Section~\ref{sec:discussion}. Second, the baselines are evaluated with a single
classifier each, while \glare\ is evaluated with five, so a fair reading of
Table~\ref{tab:main} is that a single procedure lifts a whole family of
classifiers to the level of methods that were tuned to one classifier each.
Macro-F1 scores give the same picture (Figure~\ref{fig:macro_f1}).

\begin{table}[htbp!]
  \centering
  \caption{Test accuracy (mean$\pm$std over 3 seeds) of the rewiring baselines,
  each with the classifier recommended in its own paper (in brackets), and of
  five downstream classifiers on the original graph with $\ell_2$-normalised
  features and on the \glare\ outputs. Results on the original graph with raw
  features are in Table~\ref{tab:raw}. For Tolokers, which is imbalanced, we
  also report ROC-AUC; $^{\ddagger}$ marks AUC values computed with raw features.
  The last column is the mean accuracy over the six datasets (AUC not
  included). Daggers mark the seven cells where \glare\ lowers the accuracy of
  a classifier. The top three values per column among the 15 configurations
  (summary rows not ranked) are colour-coded in bold:
  \textcolor{deepred}{\textbf{deep red}} (1st),
  \textcolor{deepblue}{\textbf{deep blue}} (2nd), and
  \textcolor{deepgreen}{\textbf{deep green}} (3rd). Full results for \glareu\
  are in Table~\ref{tab:unsup_full}.}
  \label{tab:main}
  \footnotesize
  \setlength{\tabcolsep}{2.2pt}
    \begin{adjustbox}{max width=\textwidth}
    \begin{tabular}{ll *{8}{c}}
    \toprule
    & & & & & & & \multicolumn{2}{c}{Tolokers} & \\
    \cmidrule(lr){8-9}
    Graph & Method & Actor & Squirrel & Chameleon & Roman-emp.
    & Amazon-rat. & Acc & AUC & Mean \\
    \midrule
    \multirow{5}{*}{\shortstack[l]{Rewiring\\baselines}} & ComFy (GCN) & 0.283$\pm$0.004 & 0.384$\pm$0.002 & 0.545$\pm$0.003 & 0.355$\pm$0.000 & 0.388$\pm$0.001 & 0.782$\pm$0.002 & 0.669$\pm$0.050 & 0.456 \\
     & DHGR (GCN) & \rankthree{0.350}$\pm$0.002 & \rankone{0.626}$\pm$0.019 & \rankone{0.669}$\pm$0.002 & 0.524$\pm$0.002 & 0.369$\pm$0.001 & 0.784$\pm$0.002 & 0.704$\pm$0.003 & 0.554 \\
     & FoSR (GCN) & 0.283$\pm$0.006 & 0.385$\pm$0.002 & 0.549$\pm$0.003 & 0.356$\pm$0.001 & 0.388$\pm$0.001 & 0.782$\pm$0.001 & 0.666$\pm$0.054 & 0.457 \\
     & IDGL (GCN, joint) & 0.279$\pm$0.011 & 0.406$\pm$0.011 & 0.591$\pm$0.001 & 0.281$\pm$0.003 & 0.393$\pm$0.003 & 0.782$\pm$0.000 & 0.588$\pm$0.098 & 0.455 \\
     & LPkG (GAE+LP+GNN) & 0.227$\pm$0.007 & 0.325$\pm$0.013 & 0.485$\pm$0.016 & 0.499$\pm$0.006 & \rankthree{0.487}$\pm$0.001 & 0.788$\pm$0.001 & 0.734$\pm$0.005 & 0.469 \\
    \midrule
    \multirow{5}{*}{\shortstack[l]{Original\\$\ell_2(X)$}} & GCN & 0.308$\pm$0.004 & 0.295$\pm$0.011 & 0.427$\pm$0.013 & \rankthree{0.688}$\pm$0.000 & 0.471$\pm$0.000 & 0.786$\pm$0.002 & 0.768$\pm$0.003$^{\ddagger}$ & 0.496 \\
     & GAT & 0.333$\pm$0.001 & 0.336$\pm$0.009 & 0.490$\pm$0.010 & 0.674$\pm$0.000 & 0.437$\pm$0.002 & 0.781$\pm$0.001 & 0.694$\pm$0.109$^{\ddagger}$ & 0.509 \\
     & GraphSAGE & 0.318$\pm$0.003 & 0.371$\pm$0.001 & 0.508$\pm$0.005 & \ranktwo{0.765}$\pm$0.001 & 0.462$\pm$0.003 & 0.779$\pm$0.000 & 0.774$\pm$0.005$^{\ddagger}$ & 0.534 \\
     & H2GCN & 0.314$\pm$0.006 & 0.339$\pm$0.007 & 0.442$\pm$0.005 & \rankone{0.779}$\pm$0.005 & \ranktwo{0.521}$\pm$0.004 & \rankthree{0.796}$\pm$0.002 & \rankone{0.814}$\pm$0.005$^{\ddagger}$ & 0.532 \\
     & LINKX & 0.304$\pm$0.004 & 0.327$\pm$0.004 & 0.446$\pm$0.014 & 0.601$\pm$0.003 & \rankone{0.529}$\pm$0.003 & \rankthree{0.796}$\pm$0.008 & 0.744$\pm$0.010$^{\ddagger}$ & 0.500 \\
    \midrule
    \multirow{5}{*}{\glare} & GCN & 0.349$\pm$0.008 & \ranktwo{0.521}$\pm$0.022 & 0.627$\pm$0.019 & 0.648$\pm$0.004$^{\dagger}$ & 0.481$\pm$0.005 & \ranktwo{0.797}$\pm$0.002 & \rankthree{0.802}$\pm$0.002 & 0.571 \\
     & GAT & \ranktwo{0.353}$\pm$0.010 & \rankthree{0.520}$\pm$0.020 & 0.639$\pm$0.013 & 0.653$\pm$0.004$^{\dagger}$ & 0.485$\pm$0.000 & 0.795$\pm$0.002 & 0.797$\pm$0.002 & \rankone{0.574} \\
     & GraphSAGE & \ranktwo{0.353}$\pm$0.014 & 0.501$\pm$0.008 & 0.643$\pm$0.014 & 0.659$\pm$0.001$^{\dagger}$ & 0.482$\pm$0.002 & \rankthree{0.796}$\pm$0.001 & 0.800$\pm$0.007 & \rankthree{0.572} \\
     & H2GCN & \rankone{0.356}$\pm$0.008 & 0.478$\pm$0.020 & \rankthree{0.648}$\pm$0.008 & 0.667$\pm$0.003$^{\dagger}$ & 0.484$\pm$0.006$^{\dagger}$ & \rankone{0.800}$\pm$0.002 & \ranktwo{0.806}$\pm$0.004 & \rankthree{0.572} \\
     & LINKX & 0.343$\pm$0.009 & 0.514$\pm$0.023 & \ranktwo{0.659}$\pm$0.017 & 0.657$\pm$0.002 & 0.477$\pm$0.000$^{\dagger}$ & 0.789$\pm$0.003$^{\dagger}$ & 0.771$\pm$0.010 & \ranktwo{0.573} \\
    \midrule
    \multicolumn{2}{l}{\textbf{Original, $\ell_2(X)$ (mean)}} & 0.315 & 0.334 & 0.463 & 0.701 & 0.484 & 0.788 & 0.759 & 0.514 \\
    \multicolumn{2}{l}{\textbf{\glare\ (mean)}} & 0.351 & 0.507 & 0.643 & 0.657 & 0.482 & 0.795 & 0.795 & 0.573 \\
    \multicolumn{2}{l}{\textbf{\glareu\ (mean, no labels)}} & 0.319 & 0.505 & 0.637 & 0.656 & 0.459 & 0.793 & 0.783 & 0.562 \\
    \bottomrule
  \end{tabular}
  \end{adjustbox}
\end{table}

\subsection{Uniformity across classifiers}\label{sec:res_uniform}
\begin{figure}[htbp!]
  \centering
    \includegraphics[width=0.86\textwidth,height=0.8\textheight]{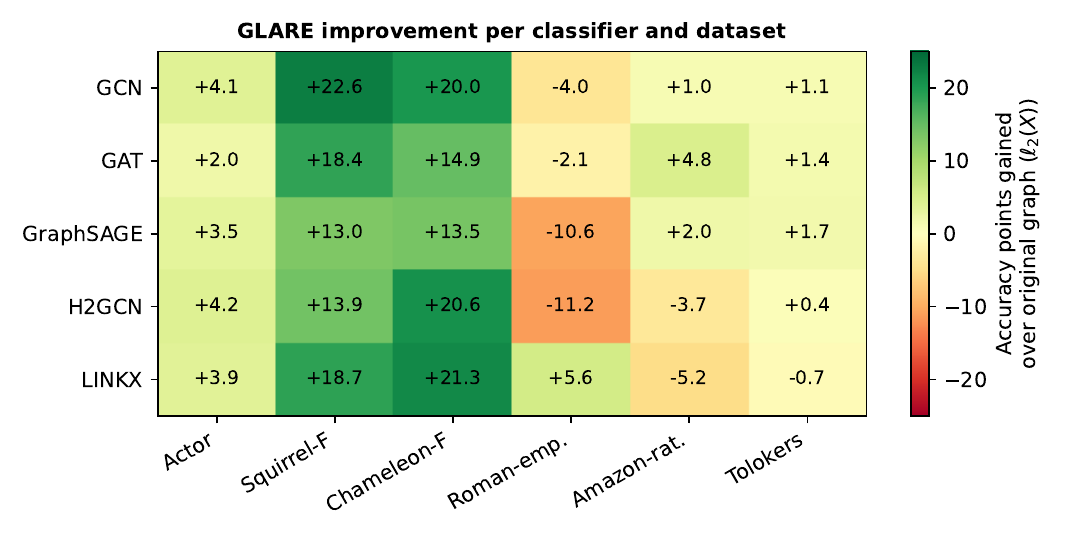}
  \caption{Accuracy points gained by \glare\ over the original graph with
  $\ell_2$-normalised features, per classifier and dataset. Green is a gain,
  red a loss. \glare\ helps in 23 of 30 cells.}
  \label{fig:heatmap}
\end{figure}

\paragraph{Improvement per classifier (RQ1).}
Comparing the original and \glare\ blocks of Table~\ref{tab:main},
\glare\ improves accuracy in 23 of 30 classifier-dataset combinations, with a
mean gain of $5.8$ points and a median of $3.7$ (Figure~\ref{fig:heatmap}). The
largest gains are on the Wikipedia graphs, for example $+22.6$ points for GCN
on Squirrel and about $+20$ points on Chameleon for GCN, H2GCN and LINKX. These
are the datasets with the lowest feature homophily, where neither the
original graph nor the raw features give a clear class signal, and where a
learned graph and embedding have the most room to help. The gains also
include the heterophily-aware classifiers: H2GCN and LINKX improve on four of
the six datasets each. The seven regressions are concentrated on
Roman-empire, where four of the five classifiers lose accuracy, and on
Amazon-ratings (H2GCN and LINKX); the seventh is a $0.7$-point drop for LINKX
on Tolokers. These are the datasets where the original graph with normalised
features is already strong.

\begin{figure}[htbp!]
  \centering
  \begin{subfigure}{0.49\textwidth}
    \includegraphics[width=\textwidth,height=0.8\textheight]{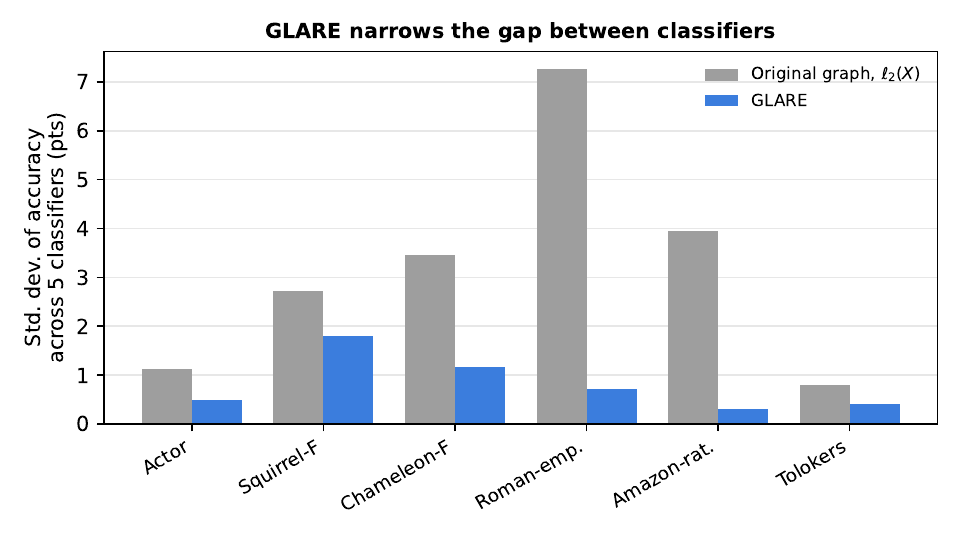}
    \caption{Spread across classifiers.}
    \label{fig:spread}
  \end{subfigure}\hfill
  \begin{subfigure}{0.49\textwidth}
    \includegraphics[width=\textwidth,height=0.8\textheight]{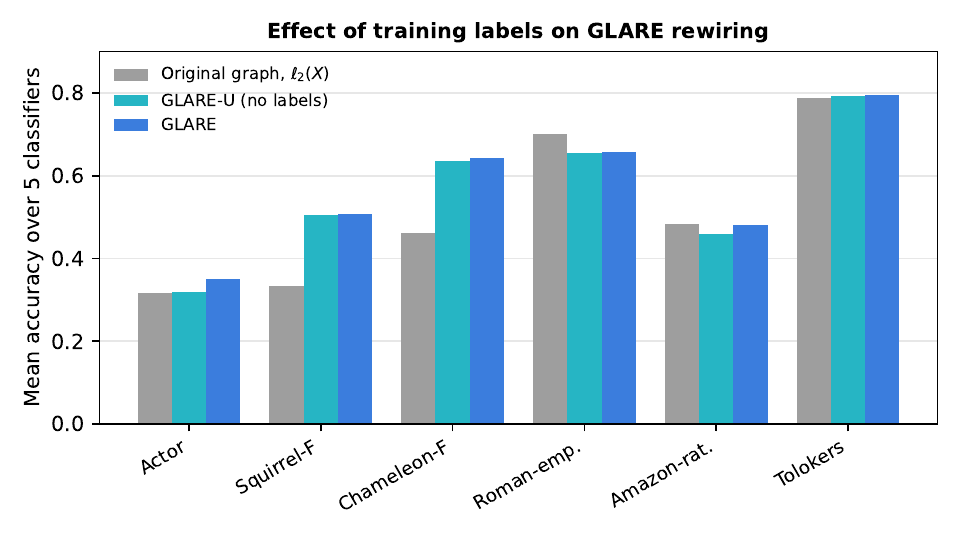}
    \caption{Effect of training labels.}
    \label{fig:supunsup}
  \end{subfigure}
  \caption{(a) \glare\ narrows the accuracy gap between the five classifiers.
  (b) Mean accuracy over the five classifiers for the original graph with
  $\ell_2(X)$, the fully unsupervised \glareu\ and \glare. \glareu\ keeps most
  of the gain without using any label during rewiring.}
\end{figure}

\paragraph{The choice of classifier matters less (RQ2).}
A practical consequence of a uniform improvement is that the downstream
classifier matters less. Figure~\ref{fig:spread} shows the standard deviation
of accuracy across the five classifiers. On the original graph this spread is
$3.2$ points on average and $7.3$ points on Roman-empire, where accuracy ranges
from $0.601$ to $0.779$ depending on the classifier. With \glare\ it falls to
$0.8$ points, and the five classifiers cluster closely on every dataset.
Simple classifiers that assume homophily, such as GCN and GAT, reach the level
of H2GCN and LINKX. Table~\ref{tab:clf_probe} provides a complementary probe:
when the learned embeddings are frozen and evaluated using graph-independent
classifiers, both Logistic Regression and a two-hidden-layer MLP achieve strong
accuracy, with \glare\ obtaining a mean of $0.559$ across the six datasets and
two probes. This indicates that useful class structure is already accessible
in the learned representation without relying on a specialized graph-based
classifier. In other words, \glare\ moves much of the burden of handling
heterophily from the architecture to its outputs, which is the main purpose of
rewiring.

\subsection{A check with ROC-AUC on Tolokers}\label{sec:res_auc}
On Tolokers a constant prediction of the majority class already gives $0.782$
accuracy, and all methods lie between about $0.78$ and $0.80$, so accuracy
hardly separates them. We therefore also report ROC-AUC, from a separate run
with the same protocol (its accuracies agree with Table~\ref{tab:main} within
$0.006$). The AUC shows clear differences. ComFy, FoSR and IDGL predict the
minority class for less than $0.5\%$ of the test nodes, so their accuracy is
essentially that of the majority class, and their AUC is only $0.59$ to $0.67$.
Every \glare\ configuration ($0.771$ to $0.806$) is above every baseline (at
most $0.734$), and \glare\ raises the AUC of four of the five classifiers over
the original graph, most for GAT ($0.694$ to $0.797$), whose AUC also becomes
much more stable across seeds. Only H2GCN is slightly lower ($0.814$ to
$0.806$). \glareu\ reaches a mean AUC of $0.783$.

\subsection{Two useful outputs: the rewired graph and the embedding}\label{sec:res_ablation}
\glare\ gives the classifier two things, the rewired graph $\tilde A$ and the
embedding $Z$ inside $\hat X=[\ell_2(X)\Vert\ell_2(Z)]$. To isolate the
contribution of each (RQ3), we train the five classifiers on four inputs: the
original or the rewired graph, combined with $\ell_2(X)$ or with $\hat X$, over
three seeds. The embedding $Z$ is always the one produced by \glare, so
``original graph with $\hat X$'' means that the embedding comes from the loop
while message passing uses the input edges. The first row of
Table~\ref{tab:ablation} is the reference of Table~\ref{tab:main}, and the last
row is a rerun of \glare\ itself ($0.570$ here against $0.573$ in
Table~\ref{tab:main}). Table~\ref{tab:ablation_full} gives the per-classifier
numbers.

\begin{table}[htbp!]
  \centering
  \caption{Rewired graph versus embedding (mean over 3 seeds). Test accuracy is
  averaged over the five downstream classifiers. ``Spread'' is the standard
  deviation of accuracy across the five classifiers, in points, averaged over
  the six datasets (lower is better). Within each two-row graph group, the
  better value is shown in bold.}
  \label{tab:ablation}
  \footnotesize
  \setlength{\tabcolsep}{2.2pt}
    \begin{adjustbox}{max width=\textwidth}
    \begin{tabular}{llcccccccc}
    \toprule
    Graph & Features & Actor & Squirrel & Chameleon & Roman-emp. &
    Amazon-rat. & Tolokers & Mean & Spread \\
    \midrule

    Original & $\ell_2(X)$
    & 0.315 & 0.334 & 0.463 & 0.701
    & 0.484 & 0.788 & 0.514 & 3.22 \\

    Original & $\hat X$
    & \textbf{0.338} & \textbf{0.492} & \textbf{0.628}
    & \textbf{0.713} & \textbf{0.501} & \textbf{0.800}
    & \textbf{0.579} & \textbf{1.78} \\

    \midrule

    Rewired & $\ell_2(X)$
    & 0.342 & 0.501 & 0.598 & 0.643
    & 0.457 & 0.785 & 0.554 & 1.47 \\

    Rewired & $\hat X$
    & \textbf{0.349} & \textbf{0.517} & \textbf{0.628}
    & \textbf{0.654} & \textbf{0.482} & \textbf{0.793}
    & \textbf{0.570} & \textbf{0.67} \\

    \bottomrule
  \end{tabular}
  \end{adjustbox}
\end{table}

We observe three things. \textbf{First, both outputs are useful on their own.}
The rewired graph alone raises the mean accuracy from $0.514$ to $0.554$
($+4.0$ points) and improves 17 of 30 cells, including all fifteen on Actor,
Squirrel and Chameleon, with gains of $+16.7$ points on Squirrel and $+13.5$ on
Chameleon. The embedding alone raises it to $0.579$ ($+6.5$ points), improves
25 of 30 cells and helps on average on all six datasets. The loop therefore
produces a useful graph and, at the same time, a useful representation.
\textbf{Second, the two outputs do different jobs.} The embedding gives the
larger accuracy gain, while the rewired graph does more to make the
classifiers agree: the spread across classifiers falls from $3.2$ points to
$1.5$ with the rewired graph alone and to $1.8$ with the embedding alone, and
it is lowest ($0.7$) when both are used. \textbf{Third, the accuracy gains
overlap.} Using both gives $0.570$, close to the embedding alone. Given
$\hat X$, the rewired graph still helps on Actor ($+1.1$) and Squirrel
($+2.5$), but it lowers accuracy on Roman-empire ($-5.9$) and Amazon-ratings
($-1.9$), where the original graph with $\hat X$ is the better choice.

A plausible explanation for the different roles is the following. The
embedding $Z$ is computed by message passing on the rewired graph, so it
already contains much of the class information that the graph provides; a
classifier that receives $Z$ gains most of the accuracy benefit whatever graph
it uses afterwards. The rewired graph, on the other hand, changes what every
message-passing classifier aggregates. Classifiers that differ mainly in how
they handle heterophilic neighbourhoods, such as GCN and H2GCN, then see a
more homophilic neighbourhood and behave more similarly, which reduces the
spread.

\begin{table}[htbp!]
  \centering
  \caption{Per-classifier test accuracy (mean$\pm$SD over 3 seeds) of the
  topology versus embedding ablation. Mean is averaged over six datasets.}
  \label{tab:ablation_full}
  \footnotesize
  \setlength{\tabcolsep}{2.2pt}
    \begin{adjustbox}{max width=\textwidth}
    \begin{tabular}{ll *{7}{c}}
    \toprule
    Input & Classifier & Actor & Squirrel & Chameleon & Roman-emp. &
    Amazon-rat. & Tolokers & Mean \\
    \midrule

    \multirow{5}{*}{\shortstack[l]{Original, $\ell_2(X)$}}
    & GCN   & 0.308{\tiny$\pm$0.004} & 0.295{\tiny$\pm$0.011} & 0.427{\tiny$\pm$0.013} & 0.688{\tiny$\pm$0.000} & 0.471{\tiny$\pm$0.000} & 0.786{\tiny$\pm$0.002} & 0.496 \\
    & GAT   & 0.333{\tiny$\pm$0.001} & 0.336{\tiny$\pm$0.009} & 0.490{\tiny$\pm$0.010} & 0.674{\tiny$\pm$0.000} & 0.437{\tiny$\pm$0.002} & 0.781{\tiny$\pm$0.001} & 0.509 \\
    & SAGE  & 0.318{\tiny$\pm$0.003} & 0.371{\tiny$\pm$0.001} & 0.508{\tiny$\pm$0.005} & 0.765{\tiny$\pm$0.001} & 0.462{\tiny$\pm$0.003} & 0.779{\tiny$\pm$0.000} & 0.534 \\
    & H2GCN & 0.314{\tiny$\pm$0.006} & 0.339{\tiny$\pm$0.007} & 0.442{\tiny$\pm$0.005} & 0.779{\tiny$\pm$0.005} & 0.521{\tiny$\pm$0.004} & 0.796{\tiny$\pm$0.002} & 0.532 \\
    & LINKX & 0.304{\tiny$\pm$0.004} & 0.327{\tiny$\pm$0.004} & 0.446{\tiny$\pm$0.014} & 0.601{\tiny$\pm$0.003} & 0.529{\tiny$\pm$0.003} & 0.796{\tiny$\pm$0.008} & 0.500 \\

    \midrule

    \multirow{5}{*}{\shortstack[l]{Original, $\hat{X}$}}
    & GCN   & 0.346{\tiny$\pm$0.004} & 0.460{\tiny$\pm$0.002} & 0.614{\tiny$\pm$0.034} & 0.721{\tiny$\pm$0.002} & 0.496{\tiny$\pm$0.005} & 0.801{\tiny$\pm$0.001} & 0.573 \\
    & GAT   & 0.347{\tiny$\pm$0.006} & 0.492{\tiny$\pm$0.018} & 0.624{\tiny$\pm$0.030} & 0.698{\tiny$\pm$0.002} & 0.497{\tiny$\pm$0.005} & 0.798{\tiny$\pm$0.004} & 0.576 \\
    & SAGE  & 0.334{\tiny$\pm$0.006} & 0.507{\tiny$\pm$0.018} & 0.639{\tiny$\pm$0.022} & 0.762{\tiny$\pm$0.004} & 0.494{\tiny$\pm$0.006} & 0.798{\tiny$\pm$0.001} & 0.589 \\
    & H2GCN & 0.336{\tiny$\pm$0.006} & 0.500{\tiny$\pm$0.020} & 0.629{\tiny$\pm$0.032} & 0.750{\tiny$\pm$0.004} & 0.494{\tiny$\pm$0.006} & 0.808{\tiny$\pm$0.006} & 0.586 \\
    & LINKX & 0.327{\tiny$\pm$0.009} & 0.501{\tiny$\pm$0.015} & 0.633{\tiny$\pm$0.032} & 0.632{\tiny$\pm$0.005} & 0.525{\tiny$\pm$0.002} & 0.794{\tiny$\pm$0.003} & 0.569 \\

    \midrule

    \multirow{5}{*}{\shortstack[l]{Rewired, $\ell_2(X)$}}
    & GCN   & 0.348{\tiny$\pm$0.004} & 0.511{\tiny$\pm$0.031} & 0.613{\tiny$\pm$0.007} & 0.637{\tiny$\pm$0.002} & 0.453{\tiny$\pm$0.002} & 0.784{\tiny$\pm$0.001} & 0.558 \\
    & GAT   & 0.346{\tiny$\pm$0.005} & 0.510{\tiny$\pm$0.012} & 0.613{\tiny$\pm$0.022} & 0.637{\tiny$\pm$0.005} & 0.421{\tiny$\pm$0.003} & 0.782{\tiny$\pm$0.001} & 0.551 \\
    & SAGE  & 0.351{\tiny$\pm$0.001} & 0.485{\tiny$\pm$0.020} & 0.598{\tiny$\pm$0.017} & 0.643{\tiny$\pm$0.004} & 0.461{\tiny$\pm$0.008} & 0.778{\tiny$\pm$0.004} & 0.553 \\
    & H2GCN & 0.337{\tiny$\pm$0.007} & 0.468{\tiny$\pm$0.019} & 0.573{\tiny$\pm$0.009} & 0.661{\tiny$\pm$0.003} & 0.479{\tiny$\pm$0.008} & 0.789{\tiny$\pm$0.005} & 0.551 \\
    & LINKX & 0.327{\tiny$\pm$0.015} & 0.532{\tiny$\pm$0.043} & 0.595{\tiny$\pm$0.003} & 0.639{\tiny$\pm$0.007} & 0.469{\tiny$\pm$0.005} & 0.790{\tiny$\pm$0.003} & 0.559 \\

    \midrule

    \multirow{5}{*}{\shortstack[l]{Rewired, $\hat{X}$}}
    & GCN   & 0.351{\tiny$\pm$0.005} & 0.531{\tiny$\pm$0.033} & 0.623{\tiny$\pm$0.017} & 0.647{\tiny$\pm$0.002} & 0.482{\tiny$\pm$0.006} & 0.797{\tiny$\pm$0.004} & 0.572 \\
    & GAT   & 0.354{\tiny$\pm$0.006} & 0.524{\tiny$\pm$0.018} & 0.628{\tiny$\pm$0.024} & 0.654{\tiny$\pm$0.002} & 0.486{\tiny$\pm$0.006} & 0.795{\tiny$\pm$0.001} & 0.573 \\
    & SAGE  & 0.350{\tiny$\pm$0.001} & 0.502{\tiny$\pm$0.025} & 0.627{\tiny$\pm$0.021} & 0.656{\tiny$\pm$0.001} & 0.478{\tiny$\pm$0.005} & 0.791{\tiny$\pm$0.006} & 0.567 \\
    & H2GCN & 0.349{\tiny$\pm$0.004} & 0.497{\tiny$\pm$0.015} & 0.626{\tiny$\pm$0.028} & 0.663{\tiny$\pm$0.003} & 0.483{\tiny$\pm$0.007} & 0.799{\tiny$\pm$0.007} & 0.569 \\
    & LINKX & 0.341{\tiny$\pm$0.010} & 0.529{\tiny$\pm$0.031} & 0.634{\tiny$\pm$0.022} & 0.652{\tiny$\pm$0.002} & 0.480{\tiny$\pm$0.006} & 0.782{\tiny$\pm$0.005} & 0.570 \\

    \bottomrule
  \end{tabular}
  \end{adjustbox}
\end{table}

\subsection{Rewiring without any labels}\label{sec:res_unsup}
Figure~\ref{fig:supunsup} compares the original graph, \glareu\ and \glare,
each as the mean over the five classifiers, and Table~\ref{tab:unsup_full}
gives the full results of \glareu. Without using a single label during
rewiring, \glareu\ improves accuracy over the original graph with $\ell_2(X)$ in
20 of 30 cells, with a mean gain of $4.8$ points, against $5.8$ for \glare
(RQ4). Training labels therefore add about one point on average, mostly on
Actor ($+3.1$) and Amazon-ratings ($+2.2$); on the other four datasets the two
variants are within one point of each other. In some cells \glareu\ is even
better than \glare, for example with GCN on Squirrel ($0.554$ against $0.521$)
and Chameleon ($0.675$ against $0.627$). The labels help more with uniformity
than with accuracy: the spread across classifiers is $2.6$ points with \glareu\
and $0.8$ with \glare, against $3.2$ on the original graph. As for \glare, the
regressions of \glareu\ are concentrated on Roman-empire and Amazon-ratings.

These results also tell us where the gain comes from. In \glareu\ the E-step
network is never trained and the homophily term is zero, so the gain of
\glareu\ comes from the contrastive encoder, the candidate pool and the
modularity objective alone. The trained E-step, which needs labels, then adds
a smaller improvement, mostly in uniformity. \glareu\ is also much cheaper:
it rewires each dataset in 56 to 158 seconds (Actor 56\,s, Squirrel 125\,s,
Chameleon 58\,s, Roman-empire 77\,s, Amazon-ratings 81\,s, Tolokers 158\,s),
because it skips the training of the E-step network and the fine-tuning of the
encoder. Table~\ref{tab:unsup_f1} reports macro-F1 scores for \glareu.

\begin{table}[htbp!]
  \centering
  \caption{Test accuracy (mean$\pm$std over 3 seeds) of the five downstream
  classifiers after fully unsupervised rewiring with \glareu, with ROC-AUC for
  Tolokers. The last column is the mean accuracy over the six datasets.}
  \label{tab:unsup_full}
  \footnotesize
  \setlength{\tabcolsep}{2.2pt}
    \begin{adjustbox}{max width=\textwidth}
    \begin{tabular}{l *{8}{c}}
    \toprule
    Classifier & Actor & Squirrel & Chameleon & Roman-emp. & Amazon-rat.
    & Tolokers & Tol.\ AUC & Mean \\
    \midrule

    GCN
    & 0.320$\pm$0.012
    & 0.554$\pm$0.006
    & 0.675$\pm$0.029
    & 0.651$\pm$0.002
    & 0.447$\pm$0.004
    & 0.792$\pm$0.005
    & 0.781$\pm$0.018
    & 0.573 \\

    GAT
    & 0.338$\pm$0.006
    & 0.502$\pm$0.003
    & 0.666$\pm$0.007
    & 0.643$\pm$0.002
    & 0.420$\pm$0.003
    & 0.788$\pm$0.004
    & 0.785$\pm$0.005
    & 0.559 \\

    SAGE
    & 0.329$\pm$0.008
    & 0.473$\pm$0.005
    & 0.645$\pm$0.023
    & 0.662$\pm$0.001
    & 0.465$\pm$0.003
    & 0.795$\pm$0.002
    & 0.794$\pm$0.003
    & 0.561 \\

    H2GCN
    & 0.317$\pm$0.015
    & 0.440$\pm$0.007
    & 0.573$\pm$0.010
    & 0.676$\pm$0.001
    & 0.491$\pm$0.009
    & 0.804$\pm$0.001
    & 0.820$\pm$0.006
    & 0.550 \\

    LINKX
    & 0.294$\pm$0.013
    & 0.557$\pm$0.018
    & 0.624$\pm$0.007
    & 0.647$\pm$0.005
    & 0.474$\pm$0.003
    & 0.788$\pm$0.002
    & 0.737$\pm$0.039
    & 0.564 \\

    \midrule

    \textbf{Mean}
    & 0.319
    & 0.505
    & 0.637
    & 0.656
    & 0.459
    & 0.793
    & 0.783
    & 0.562 \\

    \bottomrule
  \end{tabular}
  \end{adjustbox}
\end{table}

\begin{table}[htbp!]
  \centering
  \caption{Macro-F1 (mean$\pm$std over 3 seeds) of the five downstream
  classifiers after fully unsupervised rewiring with \glareu.}
  \label{tab:unsup_f1}
  \footnotesize
  \setlength{\tabcolsep}{2.2pt}
    \begin{adjustbox}{max width=\textwidth}
    \begin{tabular}{l *{7}{c}}
    \toprule
    Classifier & Actor & Squirrel & Chameleon & Roman-emp. & Amazon-rat.
    & Tolokers & Mean \\
    \midrule

    GCN
    & 0.286$\pm$0.008
    & 0.550$\pm$0.007
    & 0.674$\pm$0.030
    & 0.517$\pm$0.003
    & 0.296$\pm$0.005
    & 0.547$\pm$0.030
    & 0.478 \\

    GAT
    & 0.294$\pm$0.016
    & 0.498$\pm$0.005
    & 0.667$\pm$0.007
    & 0.492$\pm$0.003
    & 0.237$\pm$0.006
    & 0.505$\pm$0.040
    & 0.449 \\

    SAGE
    & 0.292$\pm$0.021
    & 0.465$\pm$0.006
    & 0.645$\pm$0.022
    & 0.535$\pm$0.002
    & 0.333$\pm$0.004
    & 0.607$\pm$0.018
    & 0.479 \\

    H2GCN
    & 0.278$\pm$0.022
    & 0.433$\pm$0.008
    & 0.572$\pm$0.014
    & 0.553$\pm$0.011
    & 0.427$\pm$0.009
    & 0.659$\pm$0.003
    & 0.487 \\

    LINKX
    & 0.246$\pm$0.034
    & 0.556$\pm$0.018
    & 0.625$\pm$0.008
    & 0.544$\pm$0.011
    & 0.356$\pm$0.012
    & 0.504$\pm$0.032
    & 0.472 \\

    \bottomrule
  \end{tabular}
  \end{adjustbox}
\end{table}

\subsection{Structure of the rewired graphs and embeddings}\label{sec:res_structure}
Accuracy is only one lens on a rewiring method. A graph that is genuinely
better should also be more homophilic, should support label-free graph
algorithms such as label propagation, and should reveal the class structure to
community detection. We examine these properties for \glare\ and the
baselines. All measurements in this section use the supervised \glare, and
the other methods are trained differently, so the comparisons are informative
but not fully like-for-like.

\paragraph{Homophily of the rewired graphs.}
Table~\ref{tab:dataset_grouped_delta} reports the change of the five structural
homophily measures of Section~\ref{sec:homophily_measures} after rewiring, for
\glare\ and four baselines. \glare\ gives the largest improvement on four of
the six datasets. On Roman-empire it raises adjusted homophily by $0.738$ and
label informativeness by $0.451$, and on Actor by $0.694$ and $0.481$, turning
graphs that were close to random with respect to the labels into strongly
class-consistent graphs. DHGR gives the largest improvement on Squirrel and
LPkG on Tolokers, and FoSR leaves homophily essentially unchanged, as expected
for a method that targets over-squashing. \glare\ also reduces the number of
edges on most datasets, so the higher homophily is not obtained by adding
edges. Higher homophily does not always mean higher accuracy: on Roman-empire
\glare\ gives by far the most homophilic graph but lowers the accuracy of
strong classifiers, which suggests that the removed edges carried information
that these classifiers could use even though they were heterophilic.

\begin{table}[htbp!]
\centering
\caption{%
  Comprehensive summary of structural changes among top-performing classification models, grouped by dataset. 
  We report the homophily change $\Delta$ (rewired $-$ original) across all five metrics.
  The top three performers per row are colour-coded in bold: \textcolor{deepred}{\textbf{deep red}} (1st), \textcolor{deepblue}{\textbf{deep blue}} (2nd), and \textcolor{deepgreen}{\textbf{deep green}} (3rd).
}
\label{tab:dataset_grouped_delta}
\small
\setlength{\tabcolsep}{6pt}

\begin{adjustbox}{max width=\textwidth}
\begin{tabular}{l l r r r r r}
\toprule
\textbf{Dataset} & \textbf{Metric} & \textbf{GLARE} & \textbf{DHGR} & \textbf{LPKG} & \textbf{IDGL} & \textbf{FOSR} \\
\midrule
\multirow{5}{*}{\textbf{Roman-empire}} 
  & $\Delta h_{\text{adj}}$  & \first{$+$0.738} & \second{$+$0.404} & \third{$+$0.222} & $+$0.153 & 0.000 \\
  & $\Delta h_{\text{node}}$ & \first{$+$0.609} & \second{$+$0.279} & \third{$+$0.151} & $+$0.133 & 0.000 \\
  & $\Delta h_{\text{edge}}$ & \first{$+$0.673} & \second{$+$0.389} & \third{$+$0.204} & $+$0.135 & 0.000 \\
  & $\Delta h_{\text{CI}}$   & \first{$+$0.580} & \second{$+$0.246} & \third{$+$0.151} & $+$0.070 & 0.000 \\
  & $\Delta\,\mathrm{LI}$   & \first{$+$0.451} & \second{$+$0.073} & \third{$+$0.002} & $-$0.049 & 0.000 \\
\addlinespace
\multirow{5}{*}{\textbf{Actor}} 
  & $\Delta h_{\text{adj}}$  & \first{$+$0.694} & $+$0.006 & \third{$+$0.015} & \second{$+$0.042} & $-$0.001 \\
  & $\Delta h_{\text{node}}$ & \first{$+$0.562} & \second{$+$0.035} & $+$0.006 & \third{$+$0.021} & $-$0.003 \\
  & $\Delta h_{\text{edge}}$ & \first{$+$0.548} & \third{$+$0.012} & $+$0.007 & \second{$+$0.032} & $-$0.001 \\
  & $\Delta h_{\text{CI}}$   & \first{$+$0.650} & \second{$+$0.043} & $+$0.009 & \third{$+$0.032} & $-$0.005 \\
  & $\Delta\,\mathrm{LI}$   & \first{$+$0.481} & \third{$+$0.002} & $+$0.001 & \second{$+$0.005} & 0.000 \\
\addlinespace
\multirow{5}{*}{\textbf{Squirrel}} 
  & $\Delta h_{\text{adj}}$  & \second{$+$0.193} & \first{$+$0.444} & $+$0.021 & \third{$+$0.084} & $-$0.005 \\
  & $\Delta h_{\text{node}}$ & \second{$+$0.316} & \first{$+$0.447} & $+$0.015 & \third{$+$0.072} & $+$0.009 \\
  & $\Delta h_{\text{edge}}$ & \second{$+$0.147} & \first{$+$0.341} & $+$0.008 & \third{$+$0.053} & $-$0.002 \\
  & $\Delta h_{\text{CI}}$   & \second{$+$0.179} & \first{$+$0.430} & $+$0.032 & \third{$+$0.071} & $+$0.005 \\
  & $\Delta\,\mathrm{LI}$   & \second{$+$0.051} & \first{$+$0.229} & $+$0.003 & \third{$+$0.014} & $-$0.001 \\
\addlinespace
\multirow{5}{*}{\textbf{Chameleon}} 
  & $\Delta h_{\text{adj}}$  & \first{$+$0.523} & \second{$+$0.363} & $+$0.121 & \third{$+$0.196} & $-$0.006 \\
  & $\Delta h_{\text{node}}$ & \second{$+$0.439} & \first{$+$0.474} & $+$0.080 & \third{$+$0.112} & $-$0.027 \\
  & $\Delta h_{\text{edge}}$ & \first{$+$0.418} & \second{$+$0.289} & $+$0.095 & \third{$+$0.155} & $-$0.005 \\
  & $\Delta h_{\text{CI}}$   & \first{$+$0.504} & \second{$+$0.353} & $+$0.101 & \third{$+$0.170} & $-$0.022 \\
  & $\Delta\,\mathrm{LI}$   & \first{$+$0.274} & \second{$+$0.148} & $+$0.006 & \third{$+$0.059} & $-$0.004 \\
\addlinespace
\multirow{5}{*}{\textbf{Amazon-ratings}} 
  & $\Delta h_{\text{adj}}$  & \first{$+$0.243} & \third{$+$0.065} & \second{$+$0.124} & $-$0.128 & 0.000 \\
  & $\Delta h_{\text{node}}$ & \first{$+$0.166} & \third{$+$0.070} & \second{$+$0.100} & $-$0.097 & 0.000 \\
  & $\Delta h_{\text{edge}}$ & \first{$+$0.174} & \third{$+$0.044} & \second{$+$0.084} & $-$0.100 & 0.000 \\
  & $\Delta h_{\text{CI}}$   & \first{$+$0.223} & \third{$+$0.065} & \second{$+$0.125} & $-$0.116 & 0.000 \\
  & $\Delta\,\mathrm{LI}$   & \first{$+$0.106} & \third{$+$0.011} & \second{$+$0.042} & $-$0.039 & 0.000 \\
\addlinespace
\multirow{5}{*}{\textbf{Tolokers}} 
  & $\Delta h_{\text{adj}}$  & \third{$+$0.018} & \second{$+$0.036} & \first{$+$0.140} & $-$0.056 & 0.000 \\
  & $\Delta h_{\text{node}}$ & $+$0.036 & \second{$+$0.046} & \first{$+$0.101} & \third{$+$0.039} & 0.000 \\
  & $\Delta h_{\text{edge}}$ & $+$0.018 & \third{$+$0.023} & \first{$+$0.149} & \second{$+$0.073} & 0.000 \\
  & $\Delta h_{\text{CI}}$   & $-$0.003 & \second{$+$0.014} & \first{$+$0.048} & $-$0.143 & \third{0.000} \\
  & $\Delta\,\mathrm{LI}$   & \third{$+$0.003} & \second{$+$0.006} & \first{$+$0.041} & $-$0.005 & 0.000 \\
\bottomrule
\end{tabular}
\end{adjustbox}
\end{table}

\paragraph{Label propagation.}
Label propagation \citep{zhu2002lp} assigns labels by repeatedly averaging the
label distributions of neighbours, with the training labels clamped, until
convergence. It has no learned parameters, so it measures how much the graph
alone supports the task. We run it on the original graph and on each rewired
graph and record the number of iterations and the test accuracy
(Table~\ref{tab:lp_conv_acc}). Methods that densify the graph without
filtering heterophilic edges, such as FoSR and ComFy, change the convergence
speed but end in the same uninformative, globally mixed label distribution as
the original graph. Methods that prune heterophilic edges, such as \glare\ and
DHGR, avoid this wash-out and give much higher accuracy. \glare\ has the best
mean test accuracy across datasets ($0.515$, against $0.469$ for DHGR and
$0.351$ for the original graph), with large gains on Roman-empire and
Chameleon.

\begin{table}[htbp!]
\centering
\caption{Label propagation on the \emph{rewired} graph: convergence steps and
         test-node accuracy.
         ``Steps'' is the number of LP iterations to convergence;
         ``Acc'' is the test-node majority-vote accuracy.
         The top three test-node accuracies per column are colour-coded in bold: 
         \textcolor{deepred}{\textbf{deep red}} (1st), \textcolor{deepblue}{\textbf{deep blue}} (2nd), 
         and \textcolor{deepgreen}{\textbf{deep green}} (3rd).}
\label{tab:lp_conv_acc}
\footnotesize
\setlength{\tabcolsep}{5pt}
\begin{adjustbox}{max width=\textwidth}
\begin{tabular}{l
    cc
    cc
    cc
    cc
    cc
    cc
    cc}
\toprule
& \multicolumn{2}{c}{Amazon-ratings}
& \multicolumn{2}{c}{Roman-empire}
& \multicolumn{2}{c}{Actor}
& \multicolumn{2}{c}{Chameleon}
& \multicolumn{2}{c}{Squirrel}
& \multicolumn{2}{c}{Tolokers}
& \multicolumn{2}{c}{Mean} \\
\cmidrule(lr){2-3}
\cmidrule(lr){4-5}
\cmidrule(lr){6-7}
\cmidrule(lr){8-9}
\cmidrule(lr){10-11}
\cmidrule(lr){12-13}
\cmidrule(lr){14-15}
Method
  & Steps & Acc
  & Steps & Acc
  & Steps & Acc
  & Steps & Acc
  & Steps & Acc
  & Steps & Acc
  & Steps & Acc \\
\midrule
GLARE  & 43 & \rankone{0.508} & 44 & \rankone{0.588} & 28 & 0.116         & 43 & \ranktwo{0.643} & 36 & \ranktwo{0.472} & 35 & \rankthree{0.760} & 38.2 & \rankone{0.515} \\
IDGL   & 15 & 0.328         & 27 & \rankthree{0.221} & 20 & \rankone{0.274} & 14 & \rankthree{0.478} & 12 & \rankthree{0.355} & 12 & \rankone{0.782} & 16.7 & 0.406 \\
LPKG   & 41 & \ranktwo{0.490} & 30 & \ranktwo{0.288} & 25 & \ranktwo{0.231} & 21 & 0.469         & 17 & 0.257         & 32 & \ranktwo{0.779} & 27.7 & \rankthree{0.419} \\
DHGR   & 44 & \rankthree{0.459} & 43 & 0.172         & 41 & 0.164         & 36 & \rankone{0.671} & 38 & \rankone{0.587} & 34 & 0.759         & 39.3 & \ranktwo{0.469} \\
FOSR   & 36 & 0.442         & 41 & 0.050         & 38 & 0.226         & 25 & 0.373         & 26 & 0.286         & 34 & 0.727         & 33.3 & 0.351 \\
COMFY  & 36 & 0.442         & 41 & 0.050         & 38 & \rankthree{0.228} & 35 & 0.373         & 28 & 0.290         & 34 & 0.727         & 35.3 & 0.352 \\
\midrule
\multicolumn{15}{l}{\textit{Original (baseline)}}\\
Original & 36 & 0.442         & 41 & 0.050         & 38 & 0.226         & 25 & 0.373         & 28 & 0.286         & 34 & 0.727         & 33.7 & 0.351 \\
\bottomrule
\end{tabular}
\end{adjustbox}
\end{table}

\paragraph{Community detection.}
We cluster the nodes with $k$-means, with $k$ equal to the number of classes,
on the soft labels produced by label propagation, and measure agreement with
the ground truth by normalised mutual information (NMI), adjusted Rand index
(ARI) and cluster accuracy. On the full node set (Table~\ref{tab:cd_full}),
\glare\ ranks first in mean NMI, ARI and cluster accuracy; it dominates on
Amazon-ratings and Roman-empire, while DHGR gives the best scores on Chameleon
and Squirrel. Restricting the evaluation to the test nodes, which are never
used during rewiring (Table~\ref{tab:cd_test}), gives lower scores overall, as
expected, but \glare\ again has the best means, followed by DHGR.

\begin{table}[htbp!]
\centering
\caption{Community detection on \emph{full-graph} LP-derived labels:
         NMI, ARI, and cluster accuracy (Acc) for every rewiring method and dataset.
         ``Mean'' averages each metric over all six datasets.
         The top three performers per column are colour-coded in bold: 
         \textcolor{deepred}{\textbf{deep red}} (1st), \textcolor{deepblue}{\textbf{deep blue}} (2nd), 
         and \textcolor{deepgreen}{\textbf{deep green}} (3rd).}
\label{tab:cd_full}
\footnotesize
\setlength{\tabcolsep}{2pt}
\begin{adjustbox}{max width=\textwidth}
\begin{tabular}{l
    *{3}{c}   
    *{3}{c}   
    *{3}{c}   
    *{3}{c}   
    *{3}{c}   
    *{3}{c}   
    *{3}{c}}  
\toprule
& \multicolumn{3}{c}{Amazon-ratings}
& \multicolumn{3}{c}{Roman-empire}
& \multicolumn{3}{c}{Actor}
& \multicolumn{3}{c}{Chameleon}
& \multicolumn{3}{c}{Squirrel}
& \multicolumn{3}{c}{Tolokers}
& \multicolumn{3}{c}{Mean} \\
\cmidrule(lr){2-4}\cmidrule(lr){5-7}\cmidrule(lr){8-10}
\cmidrule(lr){11-13}\cmidrule(lr){14-16}\cmidrule(lr){17-19}
\cmidrule(lr){20-22}
Method
  & NMI & ARI & Acc
  & NMI & ARI & Acc
  & NMI & ARI & Acc
  & NMI & ARI & Acc
  & NMI & ARI & Acc
  & NMI & ARI & Acc
  & NMI & ARI & Acc \\
\midrule
GLARE
  & \rankone{.400} & \rankone{.457} & \rankone{.750}
  & \rankone{.644} & \rankone{.672} & \rankone{.792}
  & \rankone{.447} & .153 & .537
  & \ranktwo{.529} & \ranktwo{.565} & \ranktwo{.799}
  & \ranktwo{.425} & \ranktwo{.420} & \ranktwo{.721}
  & .286 & .389 & .821
  & \rankone{.455} & \rankone{.443} & \rankone{.737} \\
IDGL
  & \rankthree{.382} & .256 & .523
  & .381 & \rankthree{.343} & \rankthree{.582}
  & .253 & \rankone{.269} & \rankone{.616}
  & \rankthree{.436} & \rankthree{.444} & \rankthree{.725}
  & \rankthree{.364} & \rankthree{.306} & \rankthree{.654}
  & \rankone{.445} & \ranktwo{.518} & \rankone{.891}
  & \rankthree{.377} & .356 & .665 \\
LPKG
  & \ranktwo{.391} & \ranktwo{.445} & \ranktwo{.744}
  & \rankthree{.441} & \ranktwo{.385} & \ranktwo{.596}
  & .243 & \ranktwo{.249} & \ranktwo{.606}
  & .370 & .388 & .695
  & .268 & .270 & .618
  & \ranktwo{.375} & \rankone{.554} & \ranktwo{.890}
  & .348 & \rankthree{.382} & \rankthree{.692} \\
DHGR
  & .364 & \rankthree{.423} & \rankthree{.723}
  & \ranktwo{.476} & .269 & .579
  & \ranktwo{.277} & .207 & .562
  & \rankone{.601} & \rankone{.621} & \rankone{.831}
  & \rankone{.504} & \rankone{.534} & \rankone{.782}
  & .285 & \rankthree{.442} & \rankthree{.848}
  & \ranktwo{.418} & \ranktwo{.416} & \ranktwo{.721} \\
FOSR
  & .362 & .394 & .719
  & .312 & .249 & .526
  & \rankthree{.258} & .228 & .602
  & .369 & .337 & .655
  & .353 & .214 & .606
  & \rankthree{.288} & .371 & .812
  & .324 & .299 & .654 \\
COMFY
  & .362 & .394 & .719
  & .312 & .249 & .526
  & \rankthree{.258} & \rankthree{.229} & \rankthree{.603}
  & .375 & .352 & .665
  & .355 & .215 & .607
  & .287 & .372 & .812
  & .325 & .302 & .655 \\
\bottomrule
\end{tabular}
\end{adjustbox}
\end{table}

\begin{table}[htbp!]
\centering
\caption{Community detection on \emph{test-node} LP-derived labels:
         NMI, ARI, and cluster accuracy (Acc).
         Test nodes are not used during rewiring.
         ``Mean'' averages each metric over all six datasets.
         The top three performers per column are colour-coded in bold: 
         \textcolor{deepred}{\textbf{deep red}} (1st), \textcolor{deepblue}{\textbf{deep blue}} (2nd), 
         and \textcolor{deepgreen}{\textbf{deep green}} (3rd).
         ARI is corrected for chance, so it is close to zero for an uninformative
         clustering and can be slightly negative when agreement is below chance.}
\label{tab:cd_test}
\footnotesize
\setlength{\tabcolsep}{2pt}
\begin{adjustbox}{max width=\textwidth}
\begin{tabular}{l
    *{3}{c}
    *{3}{c}
    *{3}{c}
    *{3}{c}
    *{3}{c}
    *{3}{c}
    *{3}{c}}
\toprule
& \multicolumn{3}{c}{Amazon-ratings}
& \multicolumn{3}{c}{Roman-empire}
& \multicolumn{3}{c}{Actor}
& \multicolumn{3}{c}{Chameleon}
& \multicolumn{3}{c}{Squirrel}
& \multicolumn{3}{c}{Tolokers}
& \multicolumn{3}{c}{Mean} \\
\cmidrule(lr){2-4}\cmidrule(lr){5-7}\cmidrule(lr){8-10}
\cmidrule(lr){11-13}\cmidrule(lr){14-16}\cmidrule(lr){17-19}
\cmidrule(lr){20-22}
Method
  & NMI & ARI & Acc
  & NMI & ARI & Acc
  & NMI & ARI & Acc
  & NMI & ARI & Acc
  & NMI & ARI & Acc
  & NMI & ARI & Acc
  & NMI & ARI & Acc \\
\midrule
GLARE
  & \rankone{.101} & \rankone{.105} & \ranktwo{.428}
  & \rankone{.428} & \rankone{.424} & \rankone{.573}
  & \rankone{.013} & .000 & \ranktwo{.257}
  & \ranktwo{.299} & \ranktwo{.302} & \ranktwo{.636}
  & \ranktwo{.144} & \ranktwo{.137} & \ranktwo{.466}
  & \ranktwo{.068} & \rankthree{.033} & \rankthree{.594}
  & \rankone{.175} & \rankone{.167} & \rankone{.492} \\
IDGL
  & .005 & .003 & .246
  & .089 & .040 & .173
  & \ranktwo{.012} & \rankone{.013} & \rankone{.265}
  & \rankthree{.201} & \rankthree{.166} & \rankthree{.483}
  & \rankthree{.072} & \rankthree{.057} & \rankthree{.353}
  & .030 & .012 & .559
  & .068 & .049 & .346 \\
LPKG
  & \ranktwo{.096} & \ranktwo{.095} & \rankthree{.413}
  & \ranktwo{.154} & \ranktwo{.085} & \ranktwo{.247}
  & \rankthree{.007} & \rankthree{.005} & \rankthree{.251}
  & .146 & .124 & .463
  & .012 & .006 & .264
  & \rankone{.084} & \rankone{.188} & \rankone{.747}
  & \rankthree{.083} & \rankthree{.084} & \rankthree{.397} \\
DHGR
  & \rankthree{.075} & \rankthree{.088} & \rankone{.440}
  & \rankthree{.107} & .023 & \rankthree{.243}
  & .005 & \ranktwo{.006} & .248
  & \rankone{.368} & \rankone{.324} & \rankone{.651}
  & \rankone{.248} & \rankone{.237} & \rankone{.577}
  & .052 & \ranktwo{.070} & \ranktwo{.636}
  & \ranktwo{.142} & \ranktwo{.125} & \ranktwo{.466} \\
FOSR
  & .070 & .063 & .351
  & .089 & .046 & .184
  & .004 & .001 & .233
  & .140 & .098 & .399
  & .022 & .012 & .271
  & \rankthree{.057} & .026 & .584
  & .064 & .041 & .337 \\
COMFY
  & .070 & .063 & .352
  & .085 & \rankthree{.049} & .187
  & .004 & .001 & .236
  & .142 & .100 & .397
  & .023 & .013 & .272
  & \rankthree{.057} & .027 & .585
  & .064 & .042 & .338 \\
\bottomrule
\end{tabular}
\end{adjustbox}
\end{table}

\paragraph{Quality of the embeddings.}
Methods that produce node embeddings allow two further checks. Clustering the
frozen embeddings directly (Table~\ref{tab:cd_emb}) gives \glare\ the highest
mean NMI, ARI and cluster accuracy, and the highest NMI on five of the six
datasets. Training two deliberately weak classifiers on the frozen embeddings,
a logistic regression and a two-layer perceptron (Table~\ref{tab:clf_probe}),
gives \glare\ the best mean accuracy on five of the six datasets and the best
overall mean ($0.559$, against $0.412$ for IDGL and $0.420$ for LPkG). The
class structure of the \glare\ embedding is therefore accessible even to a
linear model. Its lead is clearest on Squirrel and Chameleon, and it trails
only on Tolokers, where LPkG edges ahead. Note that the \glare\ embeddings come
from a network trained on the training labels, which favours \glare\ in this
comparison.

\begin{table}[htbp!]
\centering
\caption{Community detection in the \emph{embedding space}
         (methods that produce node embeddings only):
         NMI, ARI, and cluster accuracy (Acc).
         DHGR, FOSR, and COMFY are excluded as they do not produce embeddings.
         ``Mean'' averages each metric over all six datasets.
         The best value per column is in bold \textcolor{deepred}{\textbf{deep red}}.
         ARI is corrected for chance, so it is close to zero for an uninformative
         clustering and can be slightly negative when agreement is below chance.}
\label{tab:cd_emb}
\footnotesize
\setlength{\tabcolsep}{2pt}
\begin{adjustbox}{max width=\textwidth}
\begin{tabular}{l
    *{3}{c}
    *{3}{c}
    *{3}{c}
    *{3}{c}
    *{3}{c}
    *{3}{c}
    *{3}{c}}
\toprule
& \multicolumn{3}{c}{Amazon-ratings}
& \multicolumn{3}{c}{Roman-empire}
& \multicolumn{3}{c}{Actor}
& \multicolumn{3}{c}{Chameleon}
& \multicolumn{3}{c}{Squirrel}
& \multicolumn{3}{c}{Tolokers}
& \multicolumn{3}{c}{Mean} \\
\cmidrule(lr){2-4}\cmidrule(lr){5-7}\cmidrule(lr){8-10}
\cmidrule(lr){11-13}\cmidrule(lr){14-16}\cmidrule(lr){17-19}
\cmidrule(lr){20-22}
Method
  & NMI & ARI & Acc
  & NMI & ARI & Acc
  & NMI & ARI & Acc
  & NMI & ARI & Acc
  & NMI & ARI & Acc
  & NMI & ARI & Acc
  & NMI & ARI & Acc \\
\midrule
GLARE
  & \rankone{.030} & .003 & .249
  & \rankone{.489} & \rankone{.300} & \rankone{.403}
  & \rankone{.147} & \rankone{.114} & \rankone{.378}
  & \rankone{.532} & \rankone{.565} & \rankone{.801}
  & \rankone{.423} & \rankone{.447} & \rankone{.730}
  & .020 & $-$.002 & .527
  & \rankone{.273} & \rankone{.238} & \rankone{.515} \\
IDGL
  & .012 & \rankone{.009} & .240
  & .126 & .060 & .169
  & .031 & .032 & .312
  & .229 & .198 & .495
  & .081 & .070 & .345
  & .000 & \rankone{.000} & .500
  & .080 & .062 & .343 \\
LPKG
  & .003 & .004 & \rankone{.277}
  & .128 & .058 & .188
  & .006 & .013 & .267
  & .107 & .084 & .331
  & .004 & .002 & .222
  & \rankone{.055} & $-$.002 & \rankone{.548}
  & .051 & .027 & .306 \\
\bottomrule
\end{tabular}
\end{adjustbox}
\end{table}

\begin{table}[htbp!]
\centering
\caption{%
  Test accuracy of the three embedding-producing methods under two deliberately
  weak probes Logistic Regression (linear) and a two-hidden-layer MLP
  (shallow) trained directly on the frozen embeddings, grouped by dataset.
  Each dataset block ends with the mean over the two probes (bold); the final
  row is the mean over all 6~datasets~$\times$~2~probes. The best value per row
  is in bold \textcolor{deepred}{\textbf{deep red}}.
}
\label{tab:clf_probe}
\small
\setlength{\tabcolsep}{6pt}
\begin{adjustbox}{max width=\textwidth}
\begin{tabular}{l l r r r}
\toprule
\textbf{Dataset} & \textbf{Classifier} & \textbf{GLARE} & \textbf{IDGL} & \textbf{LPKG} \\
\midrule
\multirow{3}{*}{\textbf{Actor}}
  & Logistic Regression     & \rankone{0.356} & 0.299 & 0.274 \\
  & MLP (2-Hidden)          & \rankone{0.341} & 0.281 & 0.245 \\
  & \textbf{Mean}           & \rankone{\textbf{0.349}} & \textbf{0.290} & \textbf{0.260} \\
\midrule
\multirow{3}{*}{\textbf{Squirrel}}
  & Logistic Regression     & \rankone{0.488} & 0.367 & 0.229 \\
  & MLP (2-Hidden)          & \rankone{0.487} & 0.322 & 0.263 \\
  & \textbf{Mean}           & \rankone{\textbf{0.488}} & \textbf{0.345} & \textbf{0.246} \\
\midrule
\multirow{3}{*}{\textbf{Chameleon}}
  & Logistic Regression     & \rankone{0.640} & 0.447 & 0.336 \\
  & MLP (2-Hidden)          & \rankone{0.638} & 0.482 & 0.362 \\
  & \textbf{Mean}           & \rankone{\textbf{0.639}} & \textbf{0.465} & \textbf{0.349} \\
\midrule
\multirow{3}{*}{\textbf{Roman-empire}}
  & Logistic Regression     & \rankone{0.639} & 0.293 & 0.425 \\
  & MLP (2-Hidden)          & \rankone{0.625} & 0.240 & 0.471 \\
  & \textbf{Mean}           & \rankone{\textbf{0.632}} & \textbf{0.267} & \textbf{0.448} \\
\midrule
\multirow{3}{*}{\textbf{Amazon-ratings}}
  & Logistic Regression     & \rankone{0.475} & 0.391 & 0.401 \\
  & MLP (2-Hidden)          & \rankone{0.475} & 0.321 & 0.445 \\
  & \textbf{Mean}           & \rankone{\textbf{0.475}} & \textbf{0.356} & \textbf{0.423} \\
\midrule
\multirow{3}{*}{\textbf{Tolokers}}
  & Logistic Regression     & \rankone{0.787} & 0.782 & 0.785 \\
  & MLP (2-Hidden)          & 0.755 & 0.714 & \rankone{0.799} \\
  & \textbf{Mean}           & \textbf{0.771} & \textbf{0.748} & \rankone{\textbf{0.792}} \\
\midrule
\textbf{All datasets} & \textbf{Mean} & \rankone{\textbf{0.559}} & \textbf{0.412} & \textbf{0.420} \\
\bottomrule
\end{tabular}
\end{adjustbox}
\end{table}

\subsection{Effect of feature normalisation on the reference}\label{sec:res_raw}
All comparisons above use the original graph with $\ell_2$-normalised
features as the reference. Table~\ref{tab:raw} shows why. With raw features the
original graph reaches a mean accuracy of $0.488$ instead of $0.514$;
normalisation helps in 19 of 30 cells, and most of the difference comes from
GCN and GAT on Roman-empire, which gain $25.0$ and $26.1$ points from
normalisation alone. For GraphSAGE and H2GCN the two feature versions are within
about two points on every dataset. Against the raw-feature reference \glare\
would improve 26 of 30 cells with a mean gain of $8.4$ points. We report the
more conservative normalised comparison throughout, and we note that simple
preprocessing choices such as feature normalisation can change the
conclusions of rewiring studies and should be reported explicitly.

\begin{table}[htbp!]
  \centering
  \caption{Test accuracy (mean$\pm$std over 3 seeds) on the original graph with
  raw features $X$. The last two columns give the mean accuracy over the six
  datasets with raw features and with $\ell_2$-normalised features
  (Table~\ref{tab:main}).}
  \label{tab:raw}
  \footnotesize
  \setlength{\tabcolsep}{2.2pt}
    \begin{adjustbox}{max width=\textwidth}
    \begin{tabular}{l *{8}{c}}
    \toprule
    Classifier & Actor & Squirrel & Chameleon & Roman-emp. & Amazon-rat.
    & Tolokers & Mean ($X$) & Mean ($\ell_2(X)$) \\
    \midrule

    GCN
    & 0.276$\pm$0.010
    & 0.249$\pm$0.010
    & 0.426$\pm$0.012
    & 0.438$\pm$0.003
    & 0.464$\pm$0.004
    & 0.787$\pm$0.001
    & 0.440
    & 0.496 \\

    GAT
    & 0.274$\pm$0.007
    & 0.286$\pm$0.008
    & 0.467$\pm$0.020
    & 0.413$\pm$0.005
    & 0.436$\pm$0.001
    & 0.780$\pm$0.001
    & 0.443
    & 0.509 \\

    GraphSAGE
    & 0.320$\pm$0.006
    & 0.374$\pm$0.003
    & 0.496$\pm$0.008
    & 0.754$\pm$0.000
    & 0.462$\pm$0.001
    & 0.780$\pm$0.002
    & 0.531
    & 0.534 \\

    H2GCN
    & 0.311$\pm$0.007
    & 0.355$\pm$0.005
    & 0.440$\pm$0.012
    & 0.783$\pm$0.002
    & 0.527$\pm$0.003
    & 0.799$\pm$0.002
    & 0.536
    & 0.532 \\

    LINKX
    & 0.293$\pm$0.014
    & 0.328$\pm$0.016
    & 0.469$\pm$0.008
    & 0.555$\pm$0.004
    & 0.522$\pm$0.003
    & 0.784$\pm$0.006
    & 0.492
    & 0.500 \\

    \midrule

    \textbf{Mean}
    & 0.295
    & 0.318
    & 0.460
    & 0.589
    & 0.482
    & 0.786
    & 0.488
    & 0.514 \\

    \bottomrule
  \end{tabular}
  \end{adjustbox}
\end{table}

\section{Discussion}\label{sec:discussion}

\subsection{What drives the gains}\label{sec:disc_drivers}
Three findings stand out. First, most of the gain does not need labels:
\glareu, which uses only the contrastive encoder, the candidate pool and the
modularity objective, already gives $4.8$ of the $5.8$ points. The label-free
parts of the method are therefore responsible for most of the improvement,
and the supervised E-step mainly refines it. Second, the two outputs do
different jobs. The embedding carries most of the accuracy gain, and the
rewired graph makes the gain uniform across architectures. Third, the gains
are largest where the input is least informative. On Chameleon and Squirrel,
where both the graph and the raw features carry very little class signal
($h_{\text{adj}}$ and $h_{\text{feat}}$ below $0.04$ and $0.02$), \glare\
improves every classifier by more than ten points; on Tolokers, where the
features are strongly aligned with the edges, the gains are small.

\subsection{Practical guidance}\label{sec:disc_guidance}
Because \glare\ returns its two outputs separately, a user does not have to
take both. The results suggest a simple rule. If the downstream classifier is
fixed and strong on the original graph, as H2GCN or GraphSAGE are on
Roman-empire, the embedding alone with the original graph is the safer choice;
in Table~\ref{tab:ablation} this configuration is the best on Roman-empire
($0.713$ against $0.654$ with the rewired graph) and Amazon-ratings ($0.501$
against $0.482$). If the classifier is not yet chosen, or several classifiers
must be supported, both outputs together give the most uniform behaviour.
The choice can be made on validation data. When no labels are available at
rewiring time, \glareu\ keeps most of the benefit at a fraction of the cost.

\subsection{Relation to structure learning and joint refinement}\label{sec:disc_relation}
\glare\ sits between rewiring and graph structure learning. Like IDGL and
SLAPS, it alternates between graph and representation, but it does so in a
stand-alone stage, so the learned graph is not tied to one classifier. Our
results support this choice: the same \glare\ outputs help five classifiers,
and IDGL, whose graph is learned jointly with a GCN, is among the weakest
methods in Table~\ref{tab:main}. JDR \citep{linkerhagner2025jdr} is the method
closest in spirit, since it also returns a rewired graph and refined features
and reports that both help. The two methods differ in what they refine and
why. JDR assumes that the graph and the features are noisy views of the same
class structure and aligns their leading spectral subspaces, which makes it
well suited to graphs whose features are informative but noisy. \glare\
learns a new representation by message passing on the graph it is building
and optimises a heterophily-oriented objective, which may matter more when
the raw features are weakly aligned with the classes, as on Chameleon and
Squirrel. A direct comparison under a common protocol is needed to test this
hypothesis.

\subsection{When \glare\ does not help}\label{sec:disc_fail}
\glare\ lowers accuracy on Roman-empire for four of the five classifiers and on
Amazon-ratings for H2GCN and LINKX. In both cases the original graph with
normalised features is already strong for these classifiers, and the ablation
shows that the rewired graph, not the embedding, is responsible: the
rewiring removes edges that are heterophilic but still informative. On
Roman-empire, for example, the syntactic relations between words connect
different classes in regular ways, which heterophily-aware classifiers can
exploit and which a homophily-seeking objective discards. This points to a
limitation of any rewiring objective that rewards same-class edges: it
assumes that heterophilic edges are noise, while some of them carry structured
information. \glare\ also trails DHGR on Chameleon and Squirrel, where DHGR
produces more homophilic graphs.

\subsection{Limitations and threats to validity}\label{sec:limitations}
Several limitations should be kept in mind when interpreting our results.
\begin{itemize}
\item \emph{Datasets.} We use six medium-scale benchmarks. Chameleon and
      Squirrel are the original versions, which contain duplicated nodes; the
      filtered versions of \citet{platonov2023critical} would give a cleaner
      picture on these two datasets.
\item \emph{Splits and seeds.} We use one public split per dataset and three
      seeds, so the reported standard deviations reflect training randomness
      but not split variability. Differences below about one point should not
      be over-interpreted.
\item \emph{Baselines.} Baselines use the hyperparameters of their original
      papers and were not re-tuned on our splits, and we did not include JDR.
      A comparison with JDR under a common protocol is an important next step.
\item \emph{Metrics.} Accuracy is uninformative on Tolokers, which we address
      with ROC-AUC. The homophily measures in
      Section~\ref{sec:res_structure} use all labels and serve as descriptive
      statistics of the graphs.
\item \emph{Cost.} Supervised \glare\ trains a GNN in every outer iteration and
      is considerably slower than one-pass rewiring methods; \glareu\ is much
      cheaper. The nearest-neighbour search is quadratic in the number of
      nodes, which limits scaling to very large graphs.
\item \emph{Scope.} We study undirected, static graphs and transductive node
      classification.
\end{itemize}

\section{Conclusion and future work}\label{sec:conclusion}
We presented \glare, an affinity-guided method for heterophilic node
classification that alternates between learning node embeddings and
re-estimating the edges, and returns a rewired graph and an embedding that are
independent of the downstream classifier. We asked four questions and can now
answer them. A single rewiring procedure can improve a whole family of
classifiers roughly uniformly: \glare\ improves 23 of 30 classifier-dataset
combinations against a strong reference, with a mean gain of 5.8 points
(RQ1). It makes the choice of classifier matter much less, reducing the spread
of accuracy across classifiers about fourfold (RQ2). Its two outputs are each
useful on their own and play complementary roles, the embedding for accuracy
and the graph for uniformity (RQ3). And most of the benefit does not depend on
labels: the fully unsupervised \glareu\ keeps 4.8 of the 5.8 points (RQ4).
Beyond accuracy, the rewired graphs are more homophilic and support label
propagation and community detection better than those of the baselines.

These results suggest a broader lesson for the evaluation of rewiring methods.
Because rewiring is motivated as independent of the model, it should be
evaluated with several classifiers, and uniformity of improvement should be
reported alongside accuracy. Simple choices in the reference, such as feature
normalisation, should also be stated, since they can change the size of the
reported gains considerably.

Several directions remain open. The first is a broader evaluation, with the
filtered versions of Chameleon and Squirrel, all public splits, re-tuned
baselines and a direct comparison with JDR. The second is to make the
objective aware of informative heterophilic edges, for example by learning
which class pairs should be connected instead of rewarding only same-class
edges, which could remove the regressions on Roman-empire. The third is
scalability: approximate nearest-neighbour search and mini-batch training of
the E-step would make \glare\ applicable to graphs with millions of nodes.
Finally, extending the approach to directed, dynamic and heterogeneous graphs,
and to tasks such as link prediction, would test how general the idea of
estimating the graph and the representation together really is.

\section*{Declarations}

\subsection*{Competing interests}
The authors have no competing interests to declare that are relevant to the
content of this article.

\subsection*{Ethics approval and consent to participate}
Not applicable. The study uses only publicly available benchmark datasets and
involves no human participants or animals.

\subsection*{Consent for publication}
Not applicable.

\subsection*{Data availability}
All datasets are publicly available: Roman-empire, Amazon-ratings and
Tolokers from the repository of \citet{platonov2023critical}
(\url{https://github.com/yandex-research/heterophilous-graphs}), and Actor,
Chameleon and Squirrel through PyTorch Geometric \citep{fey2019pyg}. The
splits, result tables, rewired graphs and embeddings produced in this study
are released together with the code.

\subsection*{Code availability}
The implementation of \glare, the baselines, the ablations and the evaluation
protocol is available at \url{https://github.com/k01harshit/Affinity-Rewiring}.

\subsection*{Author contributions}
Conceptualization: S.B., H.K.; Methodology: H.K., S.B.; Software: H.K., S.C.,
P.S., P.K.; Formal analysis and investigation: H.K., S.C., P.S., P.K.;
Writing, original draft: H.K.; Writing, review and editing: all authors;
Supervision: S.B. All authors read and approved the final manuscript.

\bibliographystyle{plainnat}
\bibliography{references}

\end{document}